\documentclass[lettersize,journal]{IEEEtran}
\usepackage{amsmath,amsfonts}
\usepackage{algorithmic}
\usepackage{array}
\usepackage[caption=false,font=normalsize,labelfont=sf,textfont=sf]{subfig}
\usepackage{textcomp}
\usepackage{stfloats}
\usepackage{url}
\usepackage{verbatim}
\usepackage{graphicx}
\def\BibTeX{{\rm B\kern-.05em{\sc i\kern-.025em b}\kern-.08em
    T\kern-.1667em\lower.7ex\hbox{E}\kern-.125emX}}
\usepackage{balance}
\usepackage[pagebackref,breaklinks,colorlinks,allcolors=cvprblue]{hyperref}
\usepackage{multirow}
\usepackage[dvipsnames]{xcolor}
\hypersetup{
    colorlinks=true,
    linkcolor=red,
    filecolor=green,      
    urlcolor=magenta,
    citecolor=green,
}
\usepackage{algorithm}
\usepackage{algorithmic}
\usepackage{booktabs}
\begin{document}
\title{Visual and Textual Prompts in VLLMs for Enhancing Emotion Recognition}
\author{Zhifeng~Wang, \IEEEmembership{Student Member, IEEE,} Qixuan~Zhang,  Peter Zhang, Wenjia Niu, Kaihao Zhang, \\ Ramesh Sankaranarayana, Sabrina Caldwell, Tom Gedeon, \IEEEmembership{Senior Member, IEEE}
\thanks{Qixuan~Zhang, Zhifeng~Wang, Wenjia~Niu, Kaihao~Zhang, Ramesh~Sankaranarayana and Sabrina Caldwell were with the School of Computing, Australian National University (ANU) Canberra, ACT 2601, Australia. (e-mail: (qixuan.zhang, zhifeng.wang, wenjia.niu, Kaihao.Zhang, Ramesh.Sankaranarayana, Sabrina.Caldwell)@anu.edu.au)\\Peter Zhang was the senior AI scientist at the Quriosity Pty Ltd. (e-mail: info@quriosity.com.au \\Tom~Gedeon was with Human-Centric Advancements Chair in AI, Curtin University and Australian National University. (e-mail: tom.gedeon@curtin.edu.au)}}
\markboth{Journal of \LaTeX\ Class Files,~Vol.~18, No.~9, April ~2025}%
{How to Use the IEEEtran \LaTeX \ Templates}

\maketitle

\begin{abstract}
Vision Large Language Models (VLLMs) exhibit promising potential for multi-modal understanding, yet their application to video-based emotion recognition remains limited by insufficient spatial and contextual awareness. Traditional approaches, which prioritize isolated facial features, often neglect critical non-verbal cues such as body language, environmental context, and social interactions, leading to reduced robustness in real-world scenarios. To address this gap, we propose Set-of-Vision-Text Prompting (SoVTP), a novel framework that enhances zero-shot emotion recognition by integrating spatial annotations (e.g., bounding boxes, facial landmarks), physiological signals (facial action units), and contextual cues (body posture, scene dynamics, others’ emotions) into a unified prompting strategy. SoVTP preserves holistic scene information while enabling fine-grained analysis of facial muscle movements and interpersonal dynamics. Extensive experiments show that SoVTP achieves substantial improvements over existing visual prompting methods, demonstrating its effectiveness in enhancing VLLMs' video emotion recognition capabilities.
\end{abstract}


\section{Introduction}
Emotion recognition is a key area in computer vision \cite{dai2024multimodal}, enabling systems to understand and respond to human emotions in applications like virtual reality, HCI \cite{wang2024lrdif}, surveillance \cite{yuan2024surveillance,zhang2023geometric}, and mental health \cite{ye2024dep}. Robust emotion understanding requires not only analyzing facial expressions but also interpreting body language and contextual interactions—often under challenging visual conditions \cite{zhang2022enhanced,zhang2021deep}—and thus benefits from advanced restoration techniques \cite{zhang2018adversarial}. Traditional methods for emotion recognition \cite{yang2023context, nguyen2023micron, wang2024lldif, zhang2017facial} primarily focus on facial features, using handcrafted features or shallow machine learning models \cite{wang2024htnet}. While these approaches have shown success in controlled environments, they often fail in real-world settings where emotions are influenced by intricate interpersonal dynamics, environmental factors, and cultural variations. 
\begin{figure}[t]
  \centering
  \includegraphics[width = \linewidth]{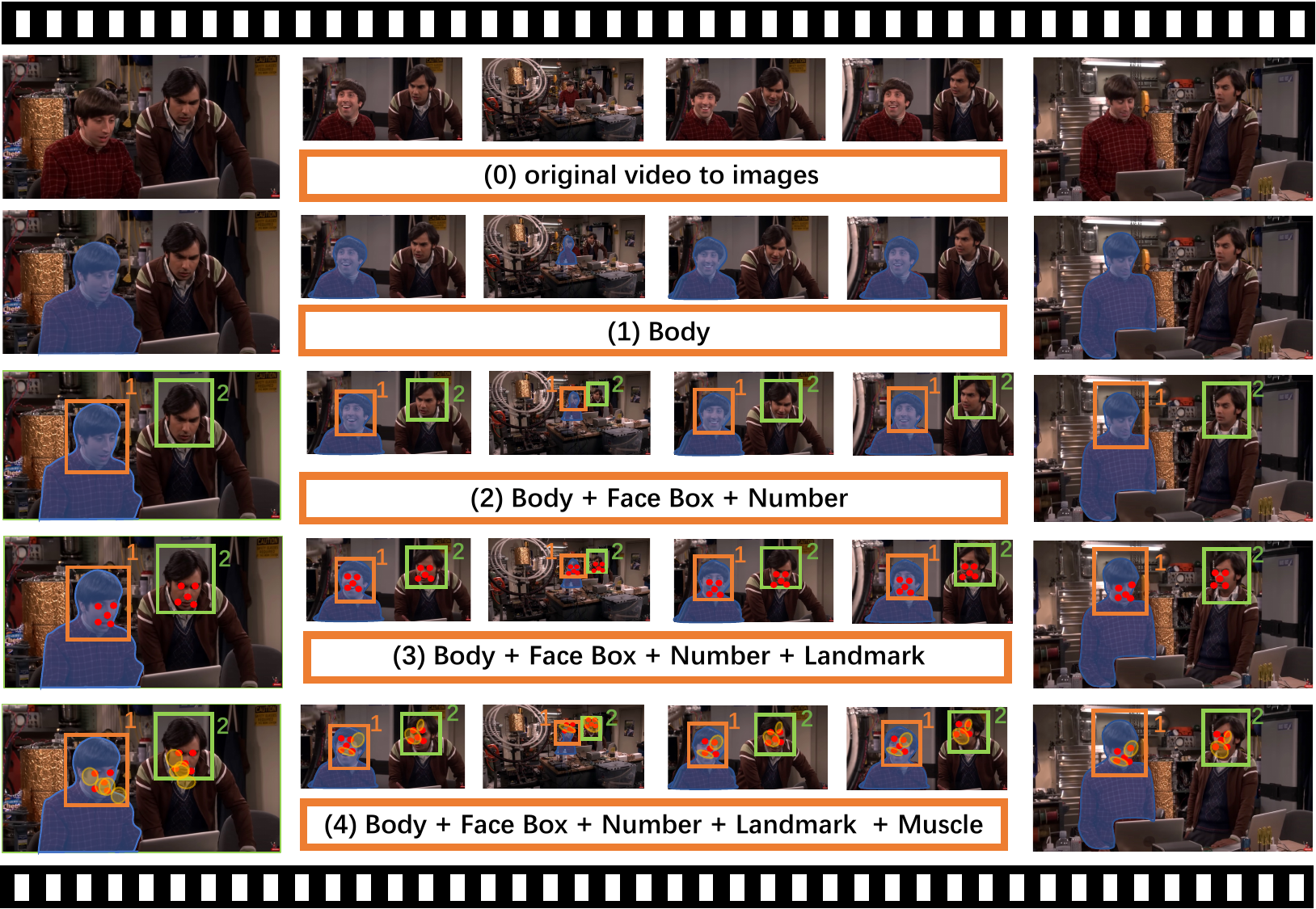} \caption{\textbf{Illustration of the SoVTP prompting for enhancing facial expression recognition in VLLMs.} The approach incrementally adds (1) body masks for person identification and tracking, (2) face boxes with numbering to ground and differentiate individuals, (3) facial landmarks to analyze and locate the spatial relationships of facial action units, and (4) facial action units for detailed muscle movement analysis. This visual prompting strategy enables fine-grained emotion recognition by preserving contextual understanding and improving model interpretability in complex scenes.
  }
  \label{introduction_four_steps_v2}
\end{figure}
\begin{figure*}[tb]
  \centering
  \includegraphics[width = \linewidth]{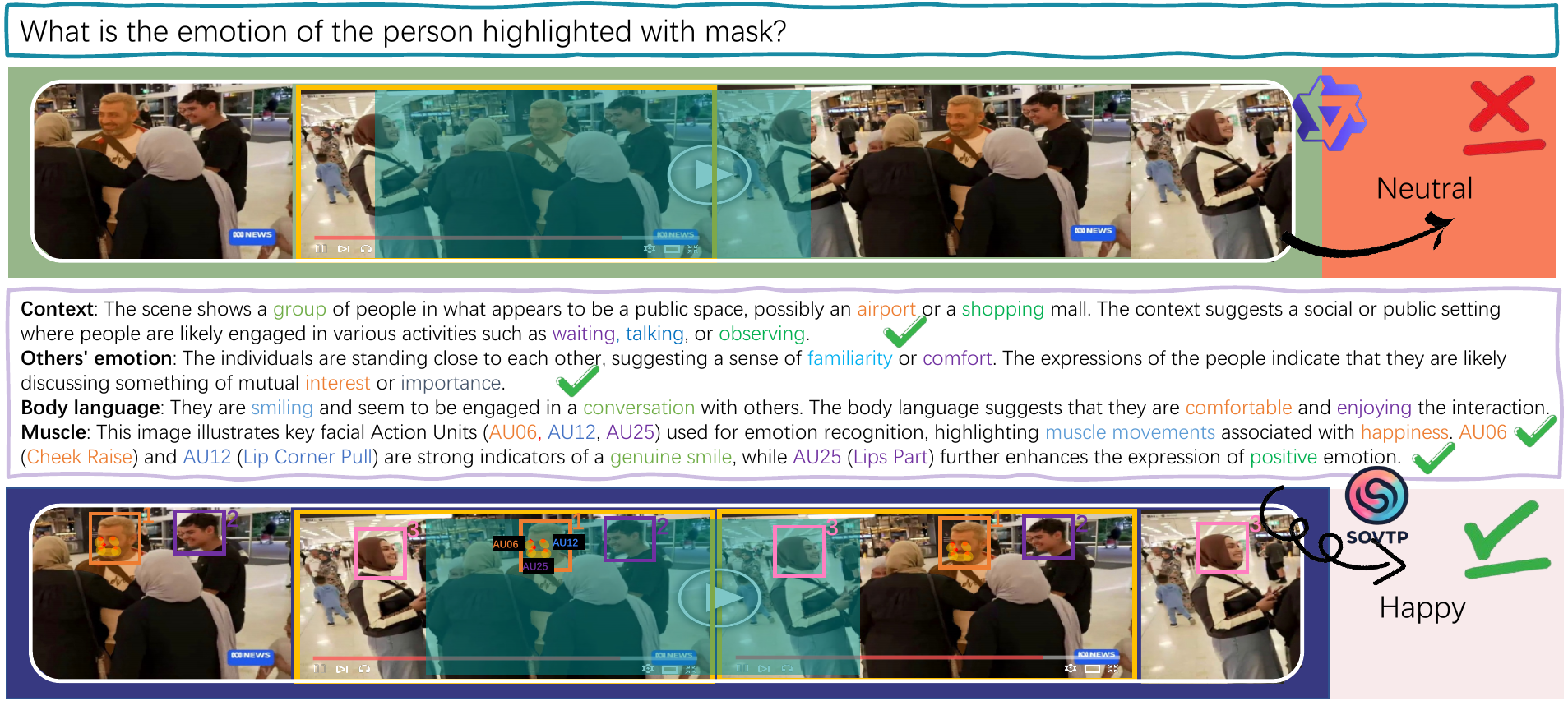}
  \caption{\textbf{Comparative analysis of emotion recognition in a complex setting}, assessing the accuracy of facial emotion classification using plain text prompts versus SoVTP prompts, which integrates context, others' emotions, body language, and facial muscle movement cues. \textbf{Top}: plain text prompts. \textbf{Bottom}: enhanced results using SoVTP. It demonstrates improved precision that incorporates environmental and social context, as well as facial muscle analysis.
  }
  \label{introduction_motivation}
\end{figure*}
Recent studies \cite{yang2024fine, zou2024segment, yang2023set} have explored zero-shot prompting techniques to enhance image-language tasks by guiding Large Language Models (LLMs) with specific visual markers. For example, RedCircle \cite{shtedritski2023does} employs red circles around objects to focus a Vision-Language Model's attention on specific regions, improving tasks like zero-shot referring expression comprehension and keypoint localization. Similarly, ReCLIP \cite{subramanian2022reclip} uses a region-scoring approach that combines colorful boxes for cropping and blurring irrelevant areas to enhance performance. Yang \textit{et al.} \cite{yang2024fine} introduced fine-grained visual prompting with segmentation masks and a `Blur Reverse Mask' technique that blurs regions outside the target area to preserve spatial context and reduce distractions. While these advances have demonstrated success in tasks like object grounding and segmentation, they face limitations when applied to emotion recognition. Methods like RedCircle and ReCLIP rely heavily on cropping techniques, which can result in the loss of critical contextual information essential for understanding emotions. Similarly, fine-grained visual prompts using Blur Reverse Mask can obscure facial features, introduce ambiguities, and hinder the accurate interpretation of nuanced emotional expressions. Although these methods hold considerable promise, their effectiveness in the domain of emotion recognition has yet to be systematically explored.

To address these limitations, this paper systematically examines various forms of visual and textual prompting. Specifically, we propose a novel framework, Set-of-Vision-Text Prompting (SoVTP), illustrated in Fig.~\ref{introduction_four_steps_v2}. SoVTP utilizes spatial information, including numerical labels, bounding boxes, facial landmarks, and action units, to precisely identify targets while preserving the broader background context. This approach significantly enhances the zero-shot performance of facial expression recognition, enabling more robust and nuanced emotion analysis.
The top part of Fig.~\ref{introduction_motivation} presents results derived from plain text prompts, indicating basic contextual information and the overall emotional assessment of individuals, but it yields incorrect results. In contrast, the bottom part of the Fig.~\ref{introduction_motivation} shows enhanced emotion recognition outcomes using SoVTP, which incorporates detailed analysis such as facial action units to identify specific muscle movements indicative of happiness. This integrated framework demonstrates improved precision in capturing nuanced emotional states by utilizing environmental, social, body language and anatomical cues, leading to more accurate differentiation between emotions such as `neutral' and `happy'. 

To summarize, our main contributions are:
\begin{itemize}
    \item The paper proposes a novel Set-of-Vision-Text Prompting (SoVTP) approach that preserves holistic scene context rather than relying on isolated facial-cropping methods by overlaying spatial annotations (e.g., face IDs, landmarks, action units) on full-scene inputs, and integrates body posture, environmental cues, and social dynamics to enrich multimodal signals.
    \item The SoVTP framework introduces a multi-stage prompting that treats zero-shot emotion recognition as structured inference over chained reasoning steps, enabling VLLMs to marginalize intermediate evidence without fine-tuning.
    \item We present comprehensive experimental results that demonstrate SoVTP significantly improves VLLMs' emotion recognition accuracy, achieving substantial gains over existing approaches.
\end{itemize}

\begin{figure*}[htb]
  \centering
  \includegraphics[width = 0.9\linewidth]{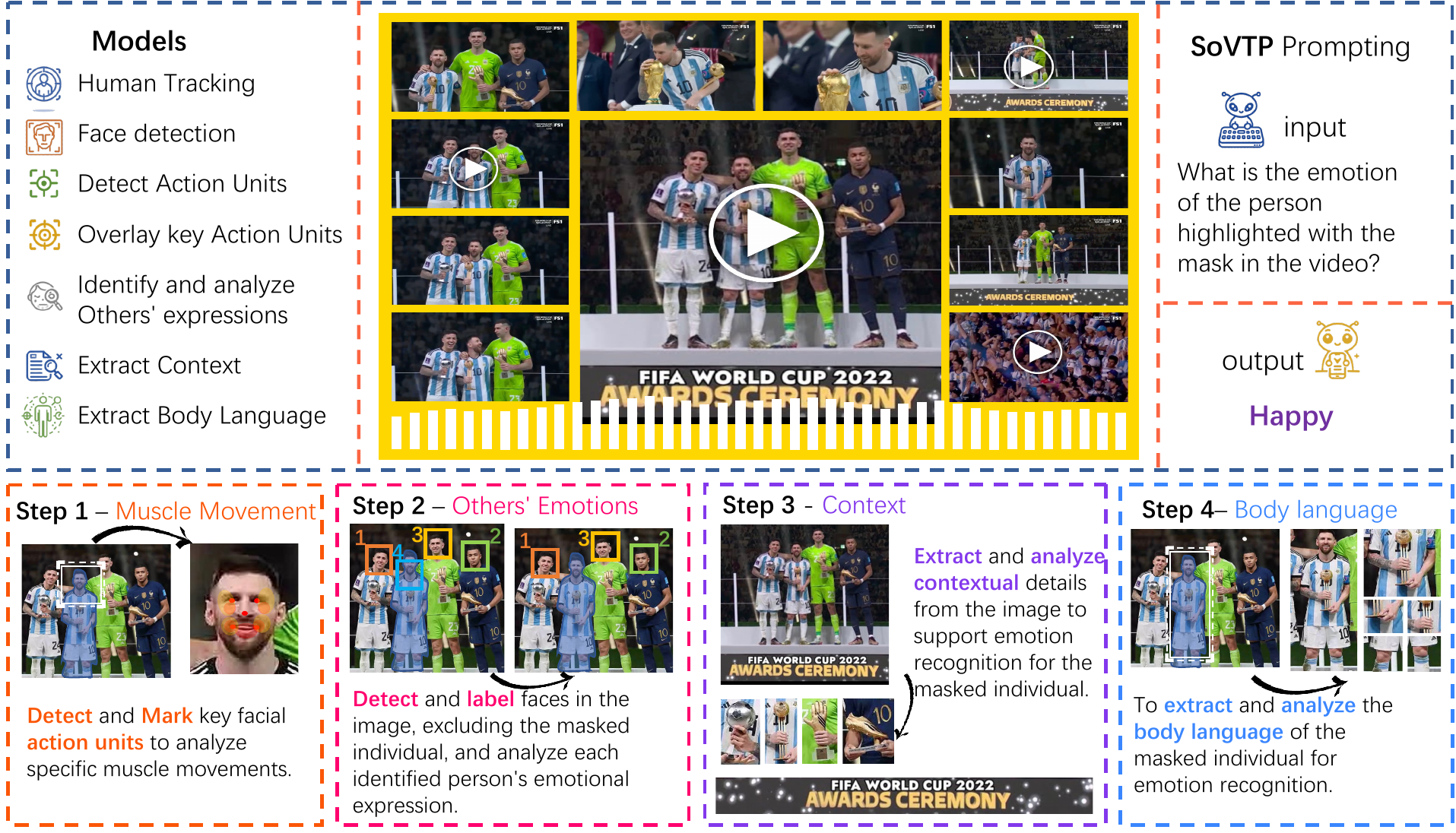}
  \caption{\textbf{Overview of the proposed SoVTP framework for emotion recognition}, illustrating a multi-step approach combining vision and text prompts. The framework includes four key steps: (1) Muscle Movement, using Action Units to analyze facial muscle activity; (2) Others' Emotions, identifying and labeling expressions of nearby individuals; (3) Context, extracting situational information to enhance emotional understanding; and (4) Body Language, assessing posture and gestures for additional emotional cues. The SoVTP prompting approach integrates these visual and contextual elements to improve accuracy in a complex scene}
  \label{method_architecture_new}
\end{figure*}
\section{Related Work}
\subsection{Vision Large Language models}
Recent advances in VLLMs have been inspired by the success of traditional LLMs such as LLaMA \cite{touvron2023llama}, PaLM \cite{chowdhery2023palm} and GPT-4 \cite{achiam2023gpt}, which have shown remarkable abilities in zero-shot generalization in a range of NLP tasks. These LLMs have demonstrated the ability to understand complex instructions and perform diverse tasks without explicit fine-tuning. The success of these models has motivated the development of VLLMs, which extend these capabilities to the visual domain by using large-scale video-text paired datasets. VideoLLaMA2 \cite{damonlpsg2024videollama2} employs an image-level CLIP encoder as the vision backbone with a novel Spatial-Temporal Convolution Connector (STC Connector) for enhanced spatial-temporal representation learning, allowing effective aggregation of frame features while preserving local details without generating excessive video tokens. Video-LLaVA \cite{lin2023video} unifies visual representations into the language feature space to create a robust VLLM, achieving significant improvements over existing models on both video and image benchmarks by leveraging mixed datasets for mutual enhancement of image and video understanding. Recent advances in action recognition and affective video analysis have laid important groundwork for emotion recognition: compositional action recognition methods that leverage progressive instance-aware feature learning \cite{yan2023progressive} and hierarchical graph-based cross-inference for group activity recognition \cite{yan2020higcin} demonstrate the value of modeling interactions and temporal dynamics, while skeleton-based techniques such as motion-aware mask feature reconstruction \cite{zhu2024motion, wang2025dm} and multi-granularity anchor-contrastive representation learning in semi-supervised settings \cite{shu2022multi} provide robust strategies for capturing fine-grained body motion cues. In parallel, decade-scale surveys of affective video content analysis \cite{xue2024affective} and frameworks like CogAware \cite{zhang2024cogaware}, transformer-based
models \cite{xenos2024vllms} or CLIP-based VLLMs \cite{zhang2023learning}  for emotion recognition, social media sentiment integration in domain-specific contexts \cite{aljedaani2022sentiment}, and unsupervised word-level affect propagation over lexical graphs \cite{fares2019unsupervised} underscore the importance of combining multimodal signals and knowledge-driven semantics. Furthermore, visual emotion analysis networks—such as object-aroused emotion analysis for images \cite{zhang2024object} and user-centered approaches integrating emotional states with visual features \cite{liang2024enhancing}—highlight effective techniques for emotion recognition. Despite these advancements, existing methods still require substantial fine-tuning of their visual and text encoders to adapt to new tasks, especially those involving nuanced visual content like emotion recognition. In contrast, our work proposes a novel zero-shot architecture for emotion recognition, aimed at using the inherent generalization abilities of VLLMs without additional training.

\subsection{Prompting methods}
Prompt engineering is a commonly used approach in NLP \cite{jiang2024delving, li2024generalizable, zhang2024visual}. ToT \cite{yao2024tree} is a novel inference framework that extends the Chain of Thought approach by enabling language models to explore multiple reasoning paths, make deliberate decisions, and evaluate alternatives, significantly enhancing performance in tasks requiring strategic planning or exploration, such as Creative Writing. Although prompts for large language models have been extensively explored, prompts for vision tasks have received less attention and investigation. Liu \textit{et al.} \cite{liu2025progressive} addresses limitations in current prompt tuning methods for vision-language models (VLMs), which use a single prompt and struggle to capture diverse visual information. While using multiple prompts could improve diversity, it risks overfitting and imbalances between base and new task performance. AnyRef \cite{he2024multi} is a general Multimodal Large Language Model (MLLM) capable of generating pixel-wise object perceptions and natural language descriptions from diverse multi-modality references, employing a refocusing mechanism to improve grounding and referring expression tasks without requiring modality-specific adaptations. FGVP \cite{yang2024fine} introduces a new zero-shot framework using pixel-level annotations from a generalist segmentation model, leveraging Blur Reverse Mask as an effective visual prompting strategy to enhance instance-level recognition tasks. While these methods have shown promising results in tasks such as semantic segmentation and object grounding, they often struggle with emotion recognition. This challenge arises from their focus on analyzing objects in isolation, neglecting essential global context and detailed facial features necessary for accurately interpreting emotional expressions. To address these shortcomings, our proposed approach emphasizes the extraction of fine-grained facial features across the entire image, maintaining spatial information through the use of spatial annotation such as boxes, numerical labels, facial landmarks, and action units. 

\section{Methods}
\subsection{Problem Definition}
The task involves matching a video to the emotions of a specific visible face within a given video sequences. This requires several steps, including human tracking, face detection, action unit analysis, context extraction, body language assessment, and emotion recognition in video sequences. Typically, VLLMs, represented as $\Theta$, process a video sequence \( V \in \mathcal{R}^{N \times H \times W \times 3} \) along with a textual query \( T^i \) as input. The model outputs an answer \( A^o \), representing the emotion, which can be expressed as (Eq. \ref{problems_definition}):
\begin{small}    
\begin{equation}
A^{o}= \Theta(V,T^{i}).
\label{problems_definition}
\end{equation}
\end{small}
Our task focuses on identifying the best matching video-emotion pairs (\(V, A^o\)) for the tracked individual. Traditionally, this process involves tracking the person and cropping their face from the image using face detection methods. However, with the proposed SoVTP prompting, it is now possible to directly highlight facial regions within the entire image, preserving the background context and avoiding the occlusion of faces. The proposed SoVTP approach overlays visual prompts onto facial regions in video sequences while simultaneously updating text prompts for VLLMs. This process transforms the original video sequence \(V\) into an enhanced sequence \(V^{new} = SoVTP(V)\), as illustrated in Fig. \ref{introduction_motivation}. The method can be formally expressed as (Eq. \ref{problems_definition_new}):
\begin{small}
\begin{equation}
A^{o}= \Theta(SoVTP(V),T^{i}).
\label{problems_definition_new}
\end{equation}
\end{small}
\begin{figure}[htbp]
  \centering
  \includegraphics[width = \linewidth]{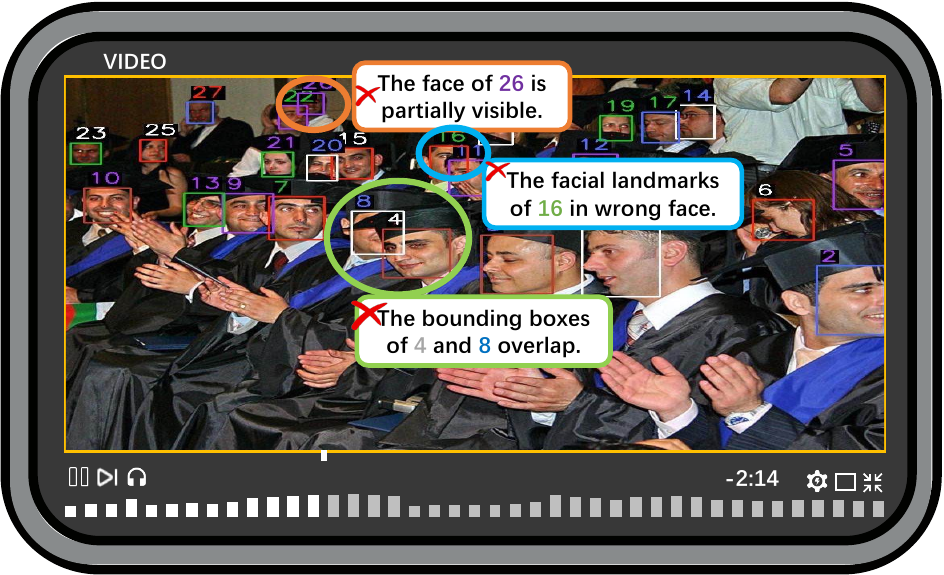}
  \caption{\textbf{Face detection often leads to overlaps or conflicts that can confuse VLLMs.} This includes challenges such as face overlaps, misaligned landmarks, and bounding box conflicts.}
  \label{overlap_images_for_algorithm}
\end{figure}
\begin{algorithm}[H]
\caption{Overlap Detection and Resolution}
\begin{small}
\begin{algorithmic}[1]
\STATE // Function to check overlap between two bounding boxes
\STATE \textbf{def} Check\_Overlap($B_1, B_2$):    
\STATE \quad $\tilde{O} = Compute\_Overlap(B_1, B_2)$ // Determine overlap
\STATE \quad \textbf{return} $\tilde{O}$
\STATE // Function to calculate the area of a bounding box
\STATE \textbf{def} Calculate\_Area(B):    
\STATE \quad $\tilde{A} = Compute\_Area(B)$ // Compute area
\STATE \quad \textbf{return} $\tilde{A}$
\STATE // Main function to handle overlapping faces
\STATE \textbf{def} Resolve\_Face\_Overlaps(faces):   
\STATE \quad // Sort detected faces in descending order of area
\STATE \quad $\tilde{F} = \textbf{Sort}(faces, \text{by = area, descending})$
\STATE \quad // Initialize a list to store visible faces
\STATE \quad visible\_faces = []
\STATE \quad // Iterate through the sorted list of faces
\STATE \quad \textbf{for} {$k$ in range(K)}:    
\STATE \quad \quad \textbf{if} Check\_Overlap($\tilde{F}[k], \tilde{F}[k+1]$):
\STATE \quad \quad \quad // Discard the smaller bounding box
\STATE \quad \quad \quad visible\_faces $= \neg \tilde{F}[k+1] \land \tilde{F}[k]$
\STATE \quad \textbf{return} visible\_faces
\end{algorithmic}
\label{overlap_solution}
\end{small}
\end{algorithm}
\subsection{Set of Vision Prompts}
\subsubsection{Box detection}
After obtaining the video sequence, the next step is to generate visual prompts that will be used by VLLMs for emotion recognition. To achieve this, we utilize the RetinaFace \cite{deng2020retinaface} algorithm to detect faces within the video sequence. Let \( B = \{b_1, b_2, \ldots, b_n\} \) represent the set of detected face bounding boxes, where \( b_i \) denotes the bounding box for the \(i\)-th face. This process can be expressed mathematically in the Eq. (\ref{box_detection}):
\begin{small}
\begin{equation}
b_{i}= \mathcal{F}(V,\alpha_{i}),
\label{box_detection}
\end{equation}
\end{small}
where \( V \) represents the video sequence, \( \alpha_{i} \) denotes the hyperparameters for the RetinaFace model \( \mathcal{F} \), and \( b_{i} \) corresponds to the bounding box of the \(i\)-th face.
\subsubsection{Face Overlap Handling Algorithm}
The face detection algorithm, while effective, often introduces overlaps or conflicts in densely populated video frames, particularly when faces overlap or one face is partially obscured by another. Such scenarios can create confusion for VLLMs, as illustrated in Fig. \ref{overlap_images_for_algorithm}. To address this issue, we propose a face overlap resolution algorithm, detailed in Algorithm \ref{overlap_solution}. 

The algorithm begins by processing a set of bounding boxes \( B = \{b_1, b_2, \ldots, b_n\} \). The area of each bounding box \( b_i \) is calculated, and the bounding boxes are sorted in descending order based on their areas (line 12) as shown in Eq. (\ref{sorted_box}):  
\begin{small}
\begin{equation}
\centering
\label{sorted_box}
B_{sorted} = \{b_1, b_2, \ldots, b_n\}, 
\end{equation}
\end{small}
where$\quad \text{Area}(b_1) \geq \text{Area}(b_2) \geq \ldots \geq \text{Area}(b_n)$, prioritizing larger bounding boxes for further processing. 
The algorithm then iterates through the sorted list \( B_{sorted} \), examining potential overlaps between the current face \( \tilde{F}[k] \) and faces already included in the final set \( F_{final} \). Overlap is evaluated using the metric in Eq. (\ref{if_overlap_box}):
\begin{small}
\begin{equation}
\begin{split}
       &\text{Overlap}(\tilde{F}[k], \tilde{F}[j]) = \\
       &\frac{\text{Area}(\tilde{F}[k] \cap \tilde{F}[j])}{\min(\text{Area}(\tilde{F}[k]), \text{Area}(\tilde{F}[j]))} > \epsilon ,
\end{split}
    \label{if_overlap_box}
\end{equation}
\end{small}
where \( \epsilon \) defines the threshold for significant overlap. $\tilde{F}[k]$ is the current face being considered for inclusion. $\tilde{F}[j]$ is an already accepted face in the final set $F_{final}$. The face $F_a$ is set to the current face $\tilde{F}[k]$ only if the overlap with the already accepted face $\tilde{F}[j]$ is less than or equal to $\epsilon$. Otherwise, $F_a$ is set to None, meaning the current face is discarded because it is considered occluded or overlaps too much, as determined by Eq. (\ref{discard_boxes}):
\begin{equation}
F_a = 
\begin{cases} 
\tilde{F}[k] & \text{if } \text{Overlap}(\tilde{F}[k], \tilde{F}[j]) <= \epsilon, \\
None & \text{otherwise}.
\end{cases}
\label{discard_boxes}
\end{equation}
Then, $F_a$ (if not None) is added to the final set $F_{final}$ using Eq. (\ref{add_final_list}):
\begin{equation}
F_{final} \leftarrow F_{final} \cup \{F_a\}.
\label{add_final_list}
\end{equation}
This process ensures that only larger faces free from obstructions are retained. After resolving overlaps and refining the bounding box set \( B = \{b_1, b_2, \ldots, b_n\} \), unique IDs \( N = \{1, 2, \ldots, n\} \) are assigned to each detected face. These IDs are then used to track faces across the video sequence, enabling distinguishable and interpretable vision prompts for VLLMs. This methodology enhances the model's ability to interpret complex visual scenes.
\begin{figure*}[htbp]
  \centering
  \includegraphics[width = \linewidth]{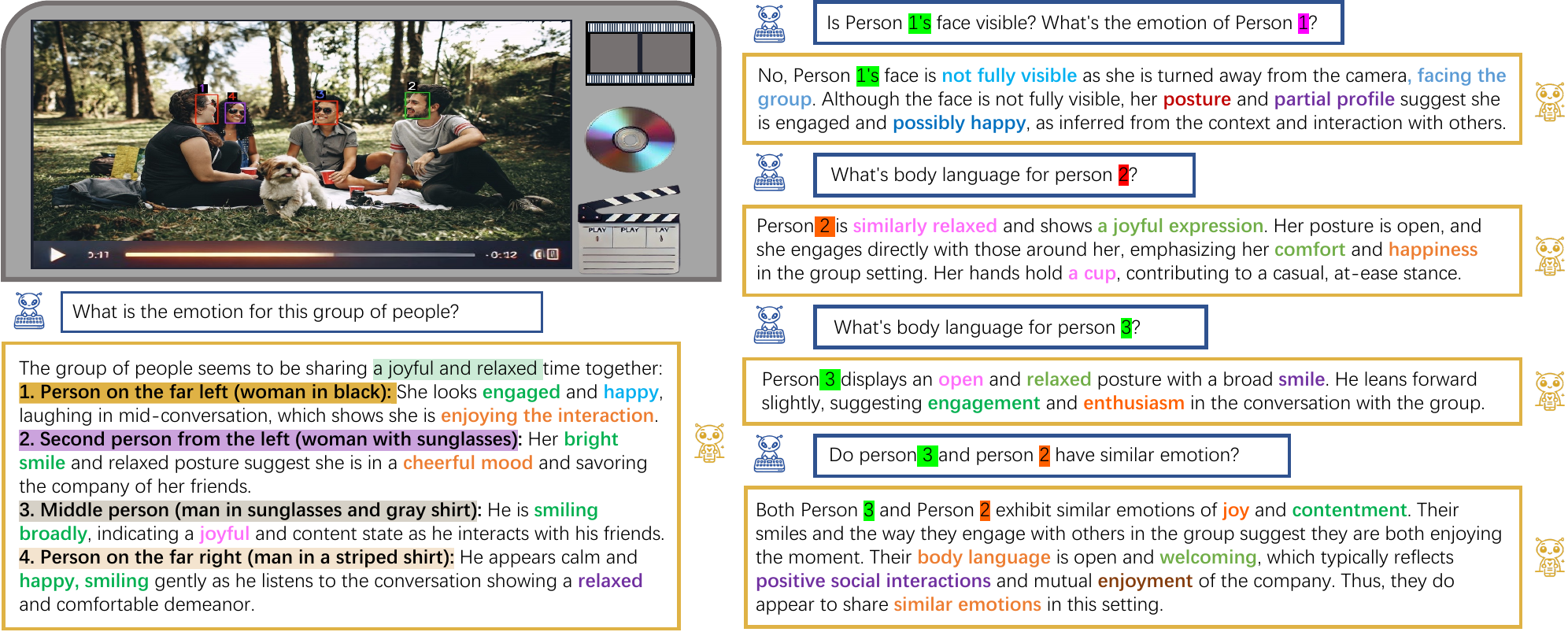}
  \caption{The figure presents two prompting approaches for emotion recognition within a social group. \textbf{Left:} It utilizes a plain-text prompt to infer group-level emotions. \textbf{Right}: It focuses on recognizing the emotions of specific individuals, as indicated by visual prompting such as numbered labels.}
  \label{insert_number_image}
\end{figure*}
\subsubsection{Muscle Movement Detection and Analysis}
The muscle movement step focuses on analyzing facial muscle activity by detecting and marking key facial action units (AUs). This process begins with the identification of the individual of interest within the scene using face detection models. Once the face is detected, we use Py-Feat \cite{cheong2023py} to extract facial landmarks. Then we analyze facial muscle deformations through action unit activation.  For detected face \( B \), the landmark detector \( L \) extracts a set of facial landmarks $M= \{l_1, l_2, \dots, l_n\}$ by Eq. (\ref{landmark_eq}):
\begin{equation}
M = L(B).
\label{landmark_eq}
\end{equation}
These landmarks correspond to action units, such as AU1 (Inner Brow Raiser), which are located at facial landmarks (20, 21, 22, 23, 24, 25). The detection model \( D \) \cite{cheong2023py} outputs the activation scores $A= \{a_1, a_2, \dots, a_m\} $ for \( m \) action units as defined in Eq. (\ref{actionunit_eq}):
\begin{equation}
A = D(\{l_1, l_2, \dots, l_n\}) .
\label{actionunit_eq}
\end{equation}
Once the activation scores are computed, they are ranked to identify the key prominent action units \( P_{ac} \), based on the initial dominant emotion extracted by Py-Feat. Here, \( P_{ac} \) is a subset of $A= \{a_1, a_2, \dots, a_m\}$ as defined in Eq. (\ref{pc_eq}):
\begin{equation}
P_{ac} = \text{Rank}(A) .
\label{pc_eq}
\end{equation}
The ranked action units \( P_{ac} \) are then overlaid onto the original face \( F \) at the positions of related facial landmarks to generate a visualization \( V \) by Eq. (\ref{visual_eq}):
\begin{equation}
V = \text{Overlay}(P_{ac}, F) .
\label{visual_eq}
\end{equation}
This step ensures that nuanced facial expressions, driven by subtle muscle movements, are captured accurately by visualization \( V \). The process is illustrated in Fig.~\ref{method_architecture_new}.
\subsection{Text and Vision Prompts}
\subsubsection{Plain Text Prompts}
The left section in Fig.~\ref{insert_number_image} utilizes a plain-text prompt to infer group-level emotions, capturing the collective sentiment of the individuals. It highlights that the group of people shares a joyful and relaxed time together, showcasing a unified emotional atmosphere. This method is effective for identifying general group emotions in shared contexts.

\subsubsection{Combined Vision-Text Prompts}
The right section in Fig.~\ref{insert_number_image} highlights the process of recognizing the emotions of specific individuals within a group setting, utilizing numbered labels to distinguish each person. For example, Person 2 and Person 3 display similar emotions, such as joy and contentment, which are inferred through their body language and interactions with others in the group. 
By capturing these interpersonal interactions and emotional exchanges, this method offers a more comprehensive analysis of social and emotional behavior within group environments.
\begin{figure}[htb]
  \centering
  \includegraphics[width = \linewidth]{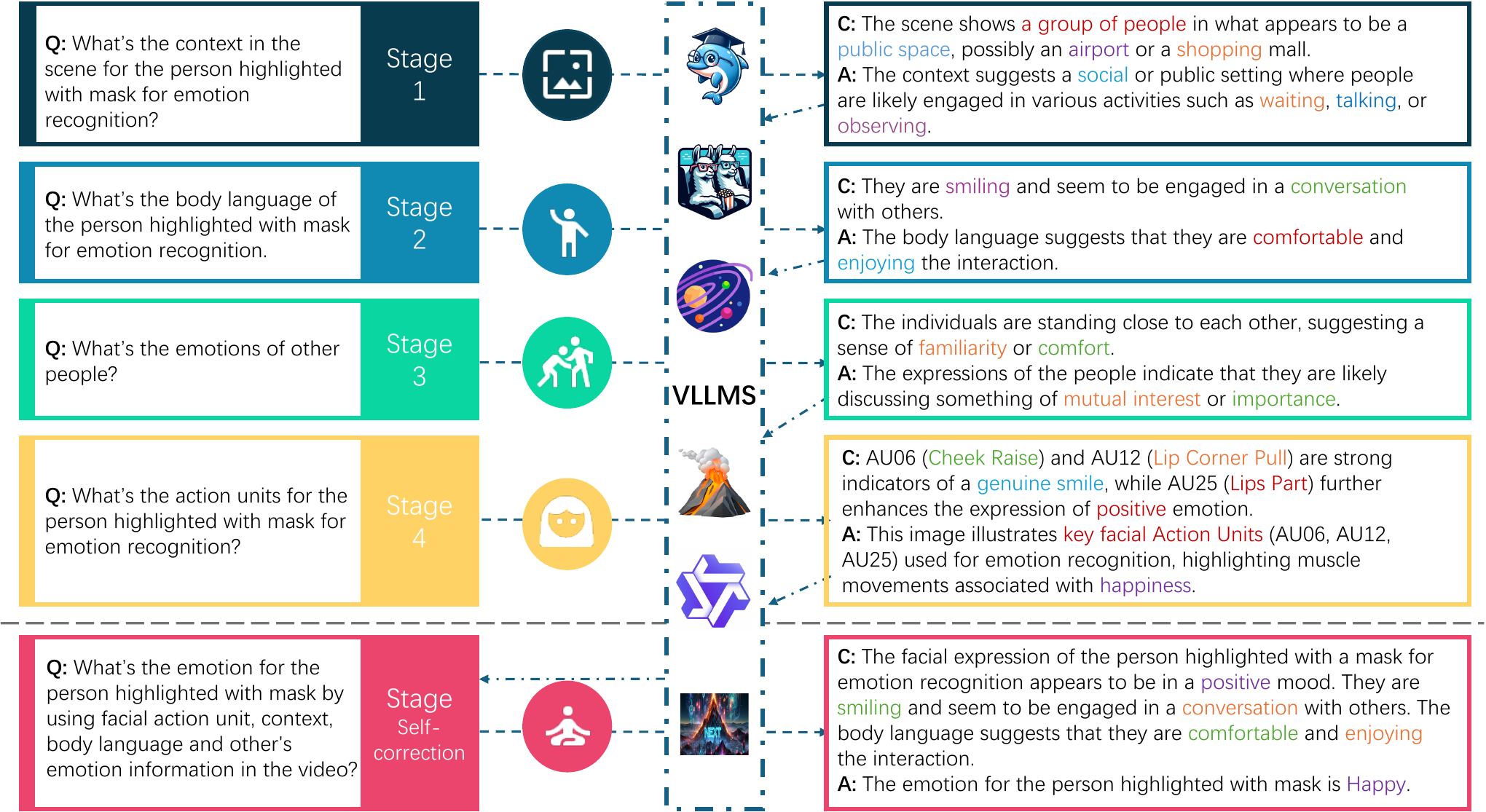}
  \caption{Overview of the multi-stage prompting strategy for emotion recognition using Vision Large Language Models (VLLMs). The approach decomposes the reasoning process into five sequential stages: (1) contextual scene understanding, (2) body language interpretation, (3) inference of emotions from surrounding individuals, (4) detection of facial action units (AUs), and (5) final emotion prediction through self-correction by integrating information from previous stages. Each stage consists of a specific question $Q_{i}$, intermediate contextual reasoning $C_{i}$, and answer $A_{i}$, facilitating a structured chain-of-thought that enhances robustness and accuracy in emotion recognition within complex social environments.}
  \label{response_text_prompting}
\end{figure}
\subsection{Final prompting strategy}
In this section, we introduce preliminaries of reasoning with VLLM prompting. For standard prompting, given the reasoning question $\mathcal{Q}$, prompt
$\mathcal{T}$, and a parameterized probabilistic vision large language model $p_{vllm}$, we aim to maximize the likelihood of an answer $\mathcal{A}$:

\begin{equation}
  p(\mathcal{A}\mid \mathcal{T},\mathcal{Q})
  = \prod_{i=1}^{|\mathcal{A}|}
    p_{vllm}\!\bigl(a_i \,\big|\, \mathcal{T},\mathcal{Q},a_{<i}\bigr).
\label{final_prompting_1}
\end{equation}

where $a_i$ is the $i$-th answer token and $|\mathcal{A}|$ is the answer length. 

For the proposed multi-stage prompting strategy that involves $K$ reasoning stages, each stage $i$ corresponds to a question-answer pair $(\mathcal{Q}_i, \mathcal{A}_i)$. In addition, we incorporate intermediate reasoning steps $\mathcal{C}_i$ chain-of-thought at each stage. The overall prompt $\mathcal{T}$ is thus composed as:
\[
\mathcal{T} = \{ (\mathcal{Q}_i, \mathcal{C}_i, \mathcal{A}_i) \}_{i=1}^K ,
\]

where

\begin{itemize}
    \item $\mathcal{Q}_i$: the question at stage $i$ (e.g., context, body language, other’s emotions, facial action units, final integration),
    \item $\mathcal{C}_i$: intermediate reasoning or evidence at stage $i$,
    \item $\mathcal{A}_i$: the answer at stage $i$.
\end{itemize}

The final answer likelihood is expressed by marginalizing over all possible reasoning trajectories $\mathcal{C} = \{ \mathcal{C}_i \}_{i=1}^K$:

\begin{equation}
p(\mathcal{A} \mid \mathcal{Q}) = \sum_{\mathcal{C}} p(\mathcal{A} \mid  \mathcal{Q}, \mathcal{C}) p(\mathcal{C} \mid  \mathcal{Q}) ,
\end{equation}
with the two factors defined by 
\begin{align}
  p(\mathcal{C}\mid \mathcal{Q})
  &= \prod_{i=1}^{|\mathcal{C}|}
     p_{\text{vllm}} (c_i \mid \mathcal{Q},c_{<i}),
  \\[6pt]
  p(\mathcal{A}\mid \mathcal{Q},\mathcal{C})
  &= \prod_{j=1}^{|\mathcal{A}|}
     p_{\text{vllm}} (a_j \mid \mathcal{Q},\mathcal{C},a_{<j}),
\end{align}
where $c_i$ denotes the $i$-th step out of the $|\mathcal{C}|$ total reasoning
steps.

In the proposed multi-stage SoVTP Prompting, Stage 1 (Context): $\mathcal{Q}_1$ asks for scene context; answer $\mathcal{A}_1$ summarizes setting.Stage 2 (Body Language): $\mathcal{Q}_2$ queries body language; $\mathcal{A}_2$ gives nonverbal cues. Stage 3 (Other’s Emotions): $\mathcal{Q}_3$ assesses surrounding individuals’ emotions; $\mathcal{A}_3$ characterizes social dynamics. Stage 4 (Facial Action Units): $\mathcal{Q}_4$ identifies key facial action units; $\mathcal{A}_4$ maps to emotional muscle activations. Stage 5 (Self-correction stage): $\mathcal{Q}_5$ integrates prior stages via $\mathcal{C}_i$ and outputs final emotion $\mathcal{A}_5$. The whole process can be described in Fig. \ref{response_text_prompting}.

The final prompting strategy combines visual and textual elements. Visual Prompts: Overlay target face annotations (numbered bounding box, facial landmarks, AUs) onto the frame, preserving background context (Fig. \ref{introduction_motivation} and Fig. \ref{method_architecture_new}). Textual Prompts include \( T_{\text{context}} \), \( T_{\text{body}} \), $T_{\text{AU}}$ and \( E_{\text{others}} \)
\begin{equation}
T^{\text{SoVTP}}_{new} = 
\left[  T_{\text{context}}; \, T_{\text{body}}; \, T_{\text{AU}}; \, E_{\text{others}} \right].
\label{final_prompting}
\end{equation}

The enhanced input \( (V^{\text{new}}, T^{\text{SoVTP}}_{new}) \) is fed into the VLLM \( \Theta \). The model analyzes and critiques body language, context, action units, and others’ emotions via chain-of-thought reasoning, then outputs the target emotion \( A^o \) as:  
\begin{equation}
A^o = \Theta(V^{\text{new}}, T^{\text{SoVTP}}_{new}).
\label{final_prompting_chain-of-thought}
\end{equation}

This integrated approach ensures the model leverages both fine-grained facial dynamics and holistic contextual information, addressing the limitations of prior methods that rely on isolated facial cropping. 

\section{Experiments}
\subsection{Models and Settings}
Our approach does not require training any models. We evaluate the proposed method's performance in a zero-shot setting using VLLMs.  Specifically, we evaluate the capabilities of several advanced VLLMs, including commercial models such as Qwen2-VL-7B-Instruct~\cite{Qwen2VL}\footnote{https://github.com/QwenLM/Qwen2-VL} and open-sourced models including ShareGPT4V-7B~\cite{chen2024sharegpt4video}\footnote{https://github.com/ShareGPT4Omni/ShareGPT4V}, VideoLLaMA2-7B \cite{damonlpsg2024videollama2}\footnote{https://github.com/DAMO-NLP-SG/VideoLLaMA2}, VILA1.5-3B \cite{lin2023vila}\footnote{https://github.com/NVlabs/VILA}, Video-LLaVA-7B-hf \cite{lin2023video}\footnote{https://github.com/PKU-YuanGroup/Video-LLaVA}, LLaVA-NeXT-Video-7B-hf \cite{zhang2024llavanext-video}\footnote{https://github.com/LLaVA-VL/LLaVA-NeXT}.  
\subsection{Dataset details} 
We collect the original videos from the ABC News website\footnote{https://www.abc.net.au/news/}. After collection, we conduct detailed preprocessing to ensure data quality such as removing duplicate and blurry videos. To reduce the effort and cost associated with manual data annotation, we utilize DeepFace \cite{serengil2024benchmark} for initial emotion labeling. These labels are then reviewed and refined by two human annotators. To ensure the accuracy of the annotations, a third annotator with expertise in psychology validate the facial expression labels based on their domain knowledge. This comprehensive process ensures high-quality annotations, which are essential for constructing reliable benchmarks. The dataset in Table \ref{dataset_introduction} for evaluating the zero-shot emotion recognition model is divided into three tiers of increasing difficulty: Easy, Medium, and Hard. The Easy tier comprises videos featuring a larger number of faces per frame, allowing the model to leverage contextual emotional cues from multiple individuals. The Medium tier includes videos with a moderate number of faces per frame, while the Hard tier contains videos with very few faces, posing a greater challenge for the model to interpret emotions due to limited social or contextual information. Across all tiers, the dataset spans a diverse collection of videos and frames, all standardized to the same resolution. It encompasses a range of emotions, including Surprise, Fear, Disgust, Anger, Happiness, Sadness, and Neutral. Performance is assessed using Accuracy and F1-Score as key metrics, ensuring a balanced evaluation of the model’s capability to detect and classify emotions. The consistent application of structured prompts across difficulty levels ensures a fair and unified testing framework, emphasizing the model’s zero-shot adaptability—its ability to generalize without prior exposure to the dataset during training. The tiered structure reflects how the number of visible faces influences task complexity, with fewer faces demanding more nuanced emotion recognition from limited visual data.
\begin{table*}[htb]
\centering
\small
\caption{\textbf{Comparison of zero-shot emotion recognition performance across various VLLMs}, including ShareGPT4V \cite{chen2024sharegpt4video}, VideoLLaMA2 \cite{damonlpsg2024videollama2}, VILA \cite{lin2023vila}, Qwen2-VL \cite{Qwen2VL}, Video-LLaVA \cite{lin2023video}, LLaVA-NeXT \cite{zhang2024llavanext-video}, and the proposed SoVTP-enhanced LLaVA-NeXT model and Qwen2-VL model. Results are reported for accuracy (Acc\%) and F1-Score (F@1) across datasets with Easy, Medium, and Hard difficulty levels.}
\resizebox{\linewidth}{!}{%
\begin{tabular}{c c c c c c c c c c c c}
\toprule
\multirow{2}{*}{\textbf{Methods}} & \multirow{2}{*}{\textbf{Backbone}} & \multicolumn{2}{c}{\textbf{Easy}} & \multicolumn{2}{c}{\textbf{Medium}} & \multicolumn{2}{c}{\textbf{Hard}} & \multicolumn{2}{c}{\textbf{Total}} \\ 
\cmidrule(lr){3-10}
& & \textbf{Acc(\%)} & \textbf{F@1} & \textbf{Acc(\%)} & \textbf{F@1}  & \textbf{Acc(\%)} & \textbf{F@1}  & \textbf{Acc(\%)} & \textbf{F@1}  \\ 
\midrule
\textbf{ShareGPT4V \cite{chen2024sharegpt4video}} & CLIP, ViT & 41.18 & 14.58 & 28.00 & 12.06 & 27.14 & 10.65 & 29.46 & 9.81 \\
\textbf{VideoLLaMA2 \cite{damonlpsg2024videollama2}} & ViT-L & 17.65 & 18.06 & \textbf{\textcolor{blue}{28.00}} & 19.05 & 27.14 & 18.35 & 25.89 & 18.28  \\
\textbf{VILA \cite{lin2023vila}} & ViT & 17.65 & 7.06 & 8.00 & 2.29 & 11.43 & 9.12 & 11.61 & 7.37 \\
\textbf{Qwen2-VL \cite{Qwen2VL}} & ViT &11.76 & 5.26 & 20.00 & 4.76 & 27.14 & 14.05 & 23.21 & 10.06 \\
\textbf{Video-LLaVA \cite{lin2023video}} & Pre-align ViT & 35.29 & 26.43 & 24.00 & 16.27 & 20.00 & 15.79 & 23.21 & 17.20\\
\textbf{LLaVA-NeXT \cite{zhang2024llavanext-video}} & SigLIP, ViT  & 29.41 & 22.92 & 12.00 & 7.91 & 17.14 & 12.45 & 17.86 & 11.05\\
\midrule
\textbf{LLaVA-NeXT \cite{zhang2024llavanext-video}+SoVTP (Ours)}  & ViT & \textbf{\textcolor{blue}{52.94}} & \textbf{\textcolor{blue}{37.38}} & 24.00 & \textbf{\textcolor{blue}{24.67}} & \textbf{\textcolor{blue}{32.86}} & \textbf{\textcolor{blue}{21.78}} & \textbf{\textcolor{blue}{33.93}} & \textbf{\textcolor{blue}{23.84}} \\
\textbf{Qwen2-VL \cite{Qwen2VL}+SoVTP (Ours)}  & ViT & \textbf{\textcolor{red}{70.59}} & \textbf{\textcolor{red}{58.00}} & \textbf{\textcolor{red}{48.00}} & \textbf{\textcolor{red}{44.94}} & \textbf{\textcolor{red}{38.57}} & \textbf{\textcolor{red}{28.25}} & \textbf{\textcolor{red}{45.54}} & \textbf{\textcolor{red}{33.53}} \\
\bottomrule
\end{tabular}
}
\label{model_performance_table}
\end{table*}

\begin{figure}[tb]
  \centering
  \includegraphics[width = \linewidth]{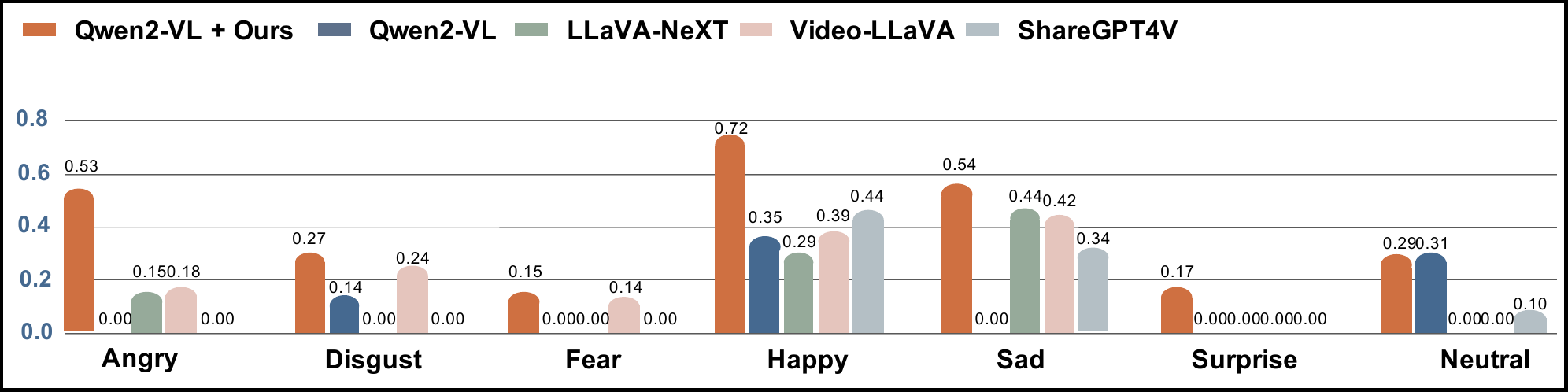}
  \caption{\textbf{Performance comparison of various VLLMs across seven emotion categories in zero-shot emotion recognition.} The bar chart depicts the accuracy of VLLMs models including Qwen2-VL+Ours, Qwen2-VL \cite{Qwen2VL}, LLaVA-NeXT \cite{zhang2024llavanext-video}, Video-LLAVA \cite{lin2023video}, and ShareGPT4V \cite{chen2024sharegpt4video}.}
  \label{compare_various_llms_results}
\end{figure}
DFEW \cite{jiang2020dfew} is a large-scale, unconstrained video database specifically designed to advance dynamic facial expression recognition in real-world conditions. It comprises 16,372 clips extracted from over 1,500 high-definition movies, each depicting one of the seven basic emotions—anger, disgust, fear, happiness, neutral, sadness, and surprise—under challenging interferences such as extreme illumination, occlusions, and pose variations. It offer a platform for the development of spatiotemporal deep learning methods for dynamic facial expression recognition in the wild.
\subsection{Action units}
Table \ref{facial_expressions_action_units} provides an overview of basic facial expressions, detailing the involved Action Units (AUs) and their descriptions. It categorizes six basic facial expressions—Surprise, Fear, Disgust, Anger, Happiness, and Sadness—each associated with specific Action Units. For Surprise, AUs 1, 2, 4, 5, 25, and 26 are involved, which include actions like Inner Brow Raiser, Outer Brow Raiser, Brow Lowerer, Upper Lid Raiser, Lip Part and Jaw Drop. Fear involves AUs 1, 2, 4, 5, 7, 11, 20, 25, and 26, incorporating actions such as Inner Brow Raiser, Outer Brow Raiser, Brow Lowerer, Upper Lid Raiser, Lid Tightener, Nasolabial Deepener, Lip Stretcher, Lip Part and Jaw Drop. Disgust involves AUs 6, 9, 11, 15, and 17, which include facial actions like Cheek Raiser, Nose Wrinkler, Nasolabial Deepener, Lip Corner Depressor and Chin Raiser. Anger is characterized by AUs 4, 5, 7, and 23, involving actions like Brow Lowerer, Upper Lid Raiser and Lid Tightener. Happiness includes AUs 6, 12, and 25, corresponding to actions like Cheek Raiser, Lip Corner Puller and Lip Part. Finally, Sadness involves AUs 1, 4, and 15, which correspond to actions like Inner Brow Raiser, Brow Lowerer and Lip Corner Depressor. This detailed mapping helps in identifying the specific facial muscle movements related to different emotional expressions.
\begin{table}[tbp]
\tiny
\centering
\caption{The table provides details about the dataset used to evaluate a model's zero-shot emotion recognition capabilities, organized into three levels of difficulty: Easy, Medium, and Hard.}
\begin{tabular}{|c|c|c|c|c|c|c|c|c|c|c|c|}
\hline
Dataset & \#Videos & \#Frames & Size &Emotions & Metrics\\
\hline
Easy & 17 & 1473 & $600\times400$ &7& Accuracy \& Recall\\
Medium & 25 & 2897 & $600\times400$ &7 & Accuracy \& Recall\\
Hard & 70 & 8276 & $600\times400$ &7 &  Accuracy \& Recall\\
Total & 112 & 12646 & $600\times400$ & 7 &  Accuracy \& Recall\\
\hline
\end{tabular}
\label{dataset_introduction}
\end{table}
\begin{table}[tbp]
    \centering
    \tiny
\caption{Basic Facial Expressions with Involved Action Units and Their Descriptions}
\begin{tabular}{|l|p{2cm}|p{3.5cm}|}
        \hline
        \textbf{Basic Expressions} & \textbf{Involved Action Units} & \textbf{Action Units Description} \\ \hline
        Surprise & AU 1, 2, 4, 5, 25, 26 & Inner Brow Raiser, Outer Brow Raiser, Brow Lowerer, Upper Lid Raiser, Lip Part, Jaw Drop \\ \hline
        Fear & AU 1, 2, 4, 5, 7, 11, 20, 25, 26 & Inner Brow Raiser, Outer Brow Raiser, Brow Lowerer, Upper Lid Raiser, Lid Tightener, Nasolabial Deepener, Lip Stretcher, Lip Part, Jaw Drop \\ \hline
        Disgust & AU 6, 9, 11, 15, 17 & Cheek Raiser, Nose Wrinkler, Nasolabial Deepener, Lip Corner Depressor, Chin Raiser \\ \hline
        Anger & AU 4, 5, 7, 23 & Brow Lowerer, Upper Lid Raiser, Lid Tightener \\ \hline
        Happiness & AU 6, 12, 25 & Cheek Raiser, Lip Corner Puller, Lip Part \\ \hline
        Sadness & AU 1, 4, 15 & Inner Brow Raiser, Brow Lowerer, Lip Corner Depressor \\ \hline
    \end{tabular}
    \label{facial_expressions_action_units}
\end{table}
\subsection{Quantitative Results}
As shown in Table~\ref{model_performance_table}, incorporating SoVTP into existing VLLMs yields substantial gains in zero‑shot emotion recognition across all difficulty tiers.  Qwen2‑VL+SoVTP achieves 70.59\% accuracy and 58.00\% F@1 on the Easy subset, 48.00\% accuracy and 44.94\% F@1 on the Medium subset, and 38.57\% accuracy and 28.25\% F@1 on the Hard subset—dramatically outperforming the strongest baselines (e.g., ShareGPT4V: 41.18\%/14.58\%, 28.00\%/12.06\%, and 27.14\%/10.65\%, respectively).  On the overall benchmark, Qwen2‑VL+SoVTP reaches 45.54\% accuracy and 33.53\% F@1, improving total accuracy by 16.08\% and F1 by 15.25\% over the best non‑SoVTP model.  LLaVA‑NeXT+SoVTP similarly boosts performance, with overall accuracy rising to 33.93\% (from 17.86\%) and F@1 to 23.84\% (from 11.05\%). These results underscore the efficacy of SoVTP in improving generalization across varying facial density contexts, with substantial gains in both recognition accuracy and robustness.
\begin{table}[tb]
\centering
\small
\caption{Comparison of zero-shot emotion recognition methods across various VLLMs on the DFEW dataset \cite{jiang2020dfew}, reporting accuracy (\%) and F@1 scores. The proposed SoVTP-enhanced Qwen2-VL and LLaVA-NeXT models demonstrate superior performance compared to existing methods.}
\resizebox{\linewidth}{!}{
\begin{tabular}{c c c c c c c c c}
\toprule
\multirow{2}{*}{\textbf{Methods}} & \multirow{2}{*}{\textbf{Backbone}} & \multirow{2}{*}{\textbf{Acc (\%)}} &\multirow{2}{*}{\textbf{F@1}}\\ \\ 
\midrule
\textbf{ShareGPT4V \cite{chen2024sharegpt4video}} & CLIP, ViT  & 25.50 & 16.85 \\ 
\textbf{VILA \cite{lin2023vila}} & ViT &24.70 & 16.22 \\
\textbf{Video-LLaVA \cite{lin2023video}} & Pre-align ViT & 28.60 & 20.78\\
\textbf{Qwen2-VL \cite{Qwen2VL}} & ViT  & 31.40 & 27.63 \\
\textbf{LLaVA-NeXT \cite{zhang2024llavanext-video}} & SigLIP, ViT  & 40.80 & 35.28\\
\midrule
\textbf{Qwen2-VL \cite{Qwen2VL}+SoVTP (Ours)}  & ViT & \textbf{\textcolor{blue}{42.30}} & \textbf{\textcolor{blue}{42.55}} \\
\textbf{LLaVA-NeXT \cite{zhang2024llavanext-video}+SoVTP (Ours)}  & ViT & \textbf{\textcolor{red}{46.50}} & \textbf{\textcolor{red}{46.79}}\\
\bottomrule
\end{tabular}
}
\label{sota_compare_dfew_dataset}
\end{table} 
\begin{figure}[tb]
  \centering
  \includegraphics[width = \linewidth]{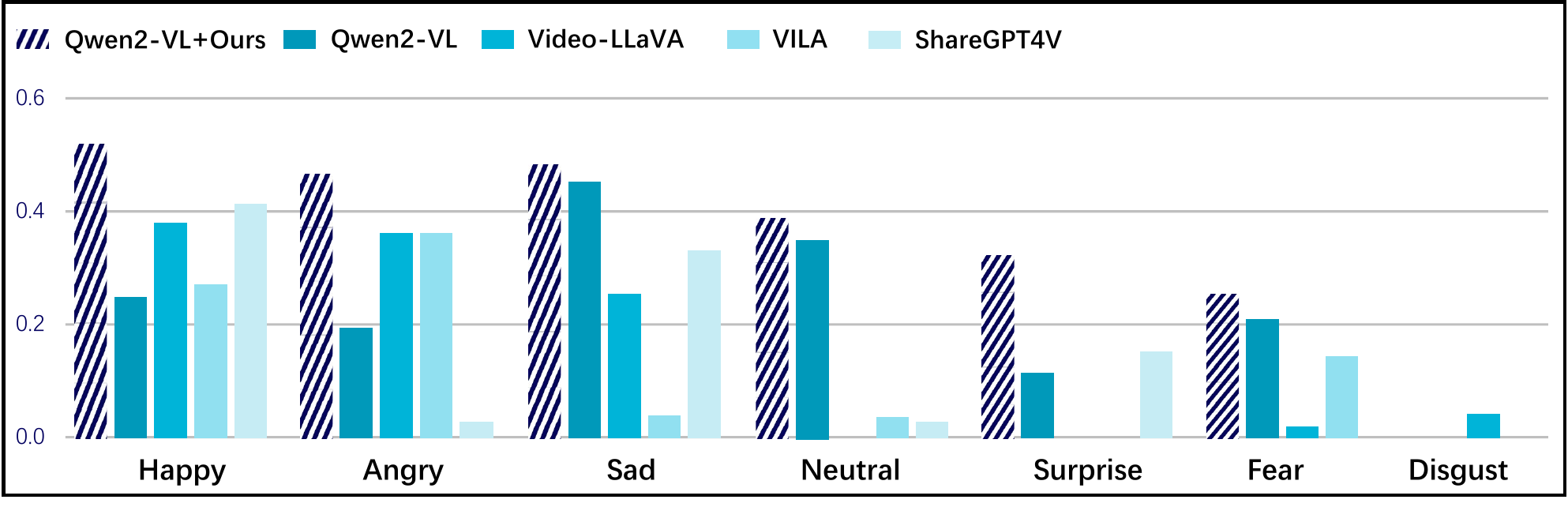}
  \caption{\textbf{Performance comparison of various VLLMs across seven emotion categories in zero-shot emotion recognition on DFEW dataset \cite{jiang2020dfew}.} The bar chart depicts the accuracy of VLLMs models including Qwen2-VL+Ours, Qwen2-VL \cite{Qwen2VL}, Video-LLAVA \cite{lin2023video}, VILA \cite{lin2023vila} and ShareGPT4V \cite{chen2024sharegpt4video}.}
  \label{dfew_compare_bar}
\end{figure}

Fig.~\ref{compare_various_llms_results} compares the zero-shot emotion recognition performance of various Vision-Language Large Models (VLLMs) across seven emotion categories: Angry, Disgust, Fear, Happy, Sad, Surprise, and Neutral. The proposed Qwen2-VL+Ours framework achieves the highest accuracy in most categories, with particularly notable performance in Happy (72\%), Surprise (54\%), and Neutral (44\%). In contrast, baseline models exhibit significant limitations: Qwen2-VL attains moderate accuracy in Happy (35\%) and Neutral (31\%) but fails to detect Angry, Fear, Sad and Surprise (0\% accuracy). LLaVA-NeXT and Video-LLaVA show marginal performance, with peak accuracies of 44\% (Sad) and 24\% (Disgust), respectively, while ShareGPT4V demonstrates minimal efficacy across all categories ($\le 10\%$ F@1). The proposed method also addresses critical gaps in underrepresented emotions, achieving 53\% accuracy in Angry, 27\% accuracy in Disgust and 54\% in Sad, where baseline models consistently score $\le$ 14\%. These results highlight the superior generalization of the Qwen2-VL+Ours framework, particularly in capturing nuanced emotional states, and underscore the limitations of existing VLLMs in handling zero-shot emotion recognition for low-frequency or complex emotions. 

Table \ref{sota_compare_dfew_dataset} summarizes the zero-shot emotion recognition performance of several vision-language models on the DFEW dataset, reporting overall accuracy and F@1 scores. Baseline methods such as ShareGPT4V, VILA, Video-LLAVA, Qwen2-VL, and LLAVA-NeXT achieve accuracies ranging from approximately 24.70\% to 40.80\% and corresponding F@1 scores of 16.22\% to 35.28\%. By contrast, the SoVTP-enhanced Qwen2-VL and LLAVA-NeXT attain substantially higher results, with Qwen2-VL+SoVTP reaching 42.30\% accuracy and 42.55\% F@1 and LLAVA-NeXT+SoVTP achieving 46.50\% accuracy and 46.79\% F@1. Fig. \ref{dfew_compare_bar} further illustrates these gains at the level of individual emotion categories (Happy, Angry, Sad, Neutral, Surprise, Fear, Disgust), showing that SoVTP-enhanced models consistently outperform their base counterparts across all categories—particularly for Happy, Angry and Surprise—thereby demonstrating the effectiveness of holistic scene prompting in improving zero-shot recognition robustness on a diverse set of emotional expressions.
\begin{table*}[ht!]
\centering
\small
\caption{\textbf{Comparison of state-of-the-art (SOTA) methods for zero-shot emotion recognition.} Previous approaches utilize various types of visual prompts: \textbf{C}: Crop, \textbf{O}: Box, \textbf{R}: Blur Reverse, \textbf{E}: Circle, \textbf{N}: Number, and \textbf{A}: Action Units.}
\resizebox{\linewidth}{!}{%
\begin{tabular}{c c c c c c c c c c}
\toprule
\multirow{2}{*}{\textbf{SOTA methods}} & \multirow{2}{*}{\textbf{Visual Prompt}} & \multicolumn{2}{c}{\textbf{Easy}} & \multicolumn{2}{c}{\textbf{Medium}} & \multicolumn{2}{c}{\textbf{Hard}} & \multicolumn{2}{c}{\textbf{Total}} \\ 
\cmidrule(lr){3-10}
& & \textbf{Acc (\%)} & \textbf{F@1} & \textbf{Acc (\%)} & \textbf{F@1} & \textbf{Acc (\%)} & \textbf{F@1} & \textbf{Acc (\%)} & \textbf{F@1} \\ 
\midrule
\textbf{Baseline \cite{Qwen2VL}} & Plain Text & 11.76 & 5.26 & 20.00 & 4.76 & 27.14 & 14.05 & 23.21 & 10.06 \\
\textbf{RedCircle \cite{shtedritski2023does}} & C $\mid$ E $\mid$ R & 29.41 & 16.62 & 36.00 & \textbf{\textcolor{blue}{33.20}} & 31.43 & 17.71 & 32.14 & 19.33 \\ 
\textbf{ReCLIP \cite{subramanian2022reclip}} & C $\mid$ O $\mid$ R & \textbf{\textcolor{blue}{35.29}} & \textbf{\textcolor{blue}{24.85}} & \textbf{\textcolor{blue}{36.00}} & 32.29 & \textbf{\textcolor{blue}{34.29}} & \textbf{\textcolor{blue}{20.75}} & \textbf{\textcolor{blue}{34.82}} & \textbf{\textcolor{blue}{23.13}} \\ 
\midrule
\textbf{SoVTP (Ours)} & N $\mid$ O $\mid$ A & \textbf{\textcolor{red}{70.59}} & \textbf{\textcolor{red}{58.00}} & \textbf{\textcolor{red}{48.00}} & \textbf{\textcolor{red}{44.94}} & \textbf{\textcolor{red}{38.57}} & \textbf{\textcolor{red}{28.25}} & \textbf{\textcolor{red}{45.54}} & \textbf{\textcolor{red}{33.53}} \\ 
\bottomrule
\end{tabular}
}
\label{sota_redclip_red_circle}
\end{table*}
\subsection{Visual Prompting}
Table \ref{sota_redclip_red_circle} presents a comparative analysis of state-of-the-art (SOTA) methods for zero-shot emotion recognition, evaluating performance across Easy, Medium, and Hard difficulty tiers using Accuracy (Acc) and F1-score (F@1) metrics. It compare the effectiveness of different visual prompting strategies utilized by Qwen2-VL. The proposed SoVTP (Ours) method, which integrates numeric annotations (N), bounding boxes (O), and facial action units (A) as visual prompts, achieves superior performance, significantly outperforming existing approaches. On the Easy tier, SoVTP attains 70.59\% Accuracy and 58.00\% F@1, a substantial improvement over the second-best method, ReCLIP (35.29\% Acc, 24.85\% F@1). Similarly, in the Medium tier, SoVTP achieves 48.00\% Accuracy and 44.94\% F@1, surpassing RedCircle (36.00\% Acc, 33.20\% F@1) and ReCLIP (36.00\% Acc, 32.29\% F@1). For the challenging Hard tier, SoVTP maintains robustness with 38.57\% Accuracy and 28.25\% F@1, outperforming ReCLIP (34.29\% Accuracy, 20.75\% F@1) and RedCircle (31.43\% Acc, 17.71\% F@1). In total metrics, SoVTP achieves 45.54\% Accuracy and 33.53\% F@1, representing a 10.72\% absolute Accuracy gain over ReCLIP (34.82\% Accuracy, 23.13\% F@1) and a 13.40\% Accuracy gain over RedCircle (32.14\% Accuracy, 19.33\% F@1). Notably, the baseline method (Qwen2-VL) with plain text prompts lags significantly across all tiers (Total: 23.21\% Accuracy, 10.06\% F@1), underscoring the critical role of structured visual prompts. The comparatively lower performance of RedCircle and ReCLIP can be attributed to their reliance on cropping and blurring techniques. This approach increases the model's focus on specific facial features but sacrifices essential body language and contextual cues, which are vital in more complex situations where emotional cues extend beyond the face. In contrast, the SoVTP approach retains a broader field of view, capturing body language and contextual information. This holistic approach accounts for the significant performance gains observed with SoVTP, especially in challenging conditions, where both facial and contextual cues are crucial for accurate emotion recognition.
\begin{figure}[htb]
  \centering
  \includegraphics[width = \linewidth]{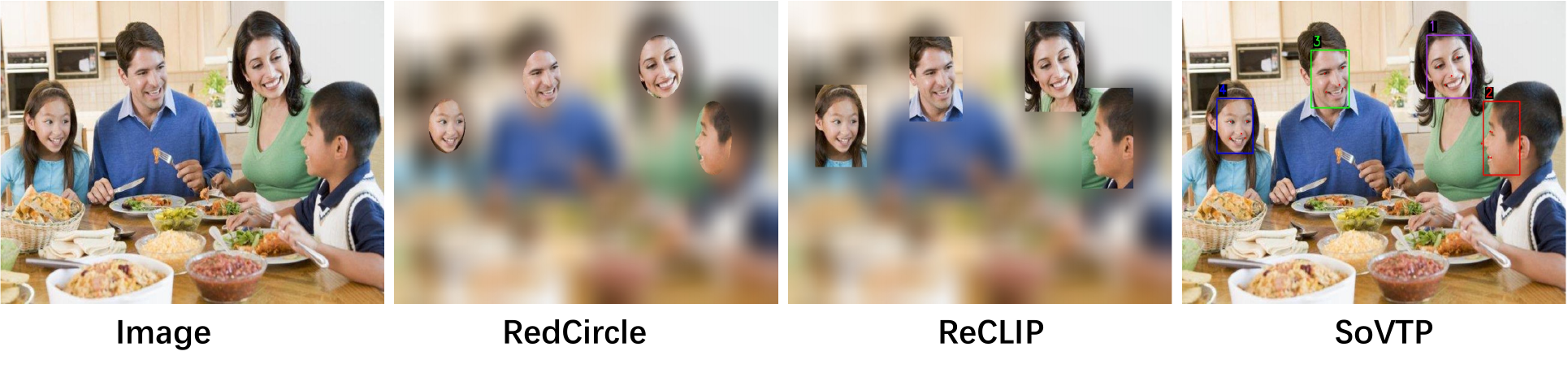}
  \caption{\textbf{Comparison of SOTA Visual Prompting Techniques in Emotion Recognition:} such as RedCircle \cite{shtedritski2023does}, ReCLIP \cite{subramanian2022reclip}, and SoVTP. RedCircle and ReCLIP blur non-face areas, focusing on facial features, while SoVTP retains the full scene, capturing body language and context, enabling a more comprehensive emotion analysis.}
  \label{sota_redcircle_clip_sovtp_experiements_label}
\end{figure}
\begin{figure}[htb]
  \centering
  \includegraphics[width = \linewidth]{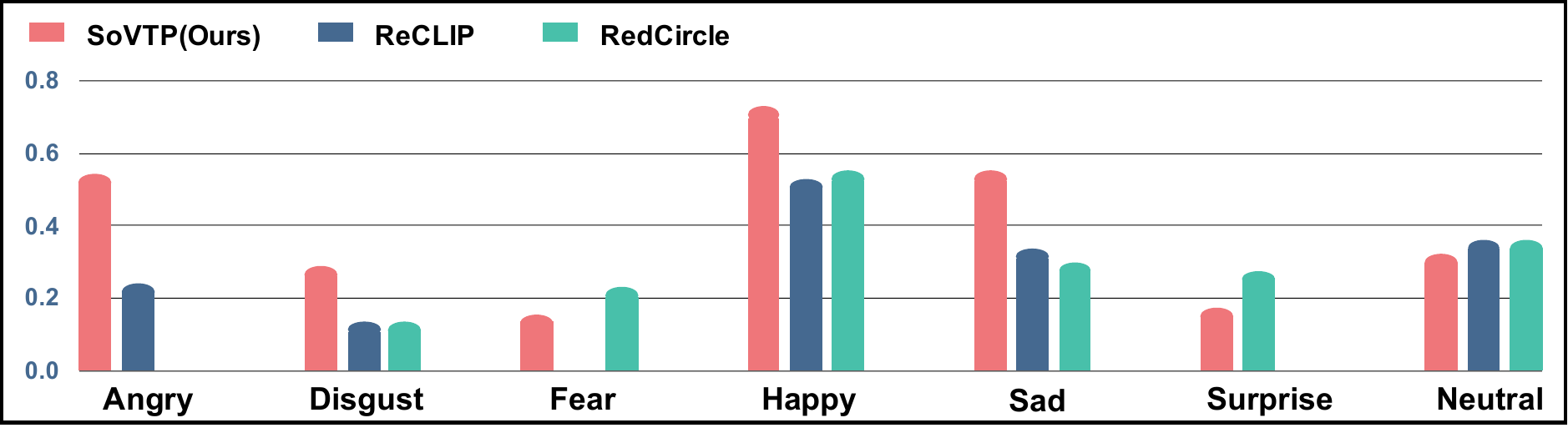}
  \caption{The bar chart displays the accuracy of different visual prompting methods such as SoVTP(Ours), ReCLIP, and RedCircle in recognizing different emotions.}
  \label{sota_compare_reclip_redcirle_bar_chart}
\end{figure}

Fig.~\ref{sota_redcircle_clip_sovtp_experiements_label} illustrates a comparative analysis of state-of-the-art (SOTA) visual prompting techniques for zero-shot emotion recognition, contrasting RedCircle, ReCLIP, and the proposed SoVTP framework. RedCircle and ReCLIP employ attention-localization strategies by blurring non-facial regions to isolate facial features (e.g., eyes, mouth), prioritizing expression-centric analysis. While this approach enhances focus on micro-expressions, it eliminates contextual cues such as body posture, hand gestures, and environmental interactions, which are critical for interpreting emotions in socially complex scenarios. In contrast, SoVTP retains the full scene without spatial or contextual obfuscation, preserving both facial features and holistic environmental information. This enables the model to integrate multi-modal cues—such as body language, spatial relationships between individuals, and scene dynamics—into its emotion recognition pipeline. For instance, the image depicts a nuclear family of four—mother, father, daughter and son—seated around a well‑lit kitchen dining table laden with a variety of home‑cooked dishes. The parents sit opposite each other, smiling warmly as they engage with their children: the mother leans in attentively toward her son, who appears to be speaking or laughing, while the father holds a fork mid‑air, looking toward his daughter with a gentle, encouraging expression. The daughter, seated at the table’s end, gazes at her parents with an open, cheerful countenance. The overall scene conveys a relaxed, intimate mealtime atmosphere, with each family member exhibiting positive affect—smiles, eye contact, and open body language—suggesting feelings of warmth, belonging, and mutual enjoyment. However, the contextual cues are obscured in RedCircle and ReCLIP, they remain visible in SoVTP, allowing the model to infer emotions like happiness or mutual enjoyment more accurately. The figure underscores SoVTP’s methodological advantage in scenarios requiring contextual awareness (e.g., group interactions, ambiguous expressions), where auxiliary non-verbal signals significantly augment recognition robustness. Quantitative improvements observed in prior experiments (e.g., 45.54\% total accuracy for SoVTP vs. $\le$ 34.82\% for SOTA baselines) align with this qualitative analysis, validating the importance of scene preservation for generalizable emotion inference. The comparison highlights a paradigm shift from isolated facial analysis to context-aware modeling, addressing a critical limitation in existing visual prompting frameworks.

Fig.~\ref{sota_compare_reclip_redcirle_bar_chart} presents a comparative bar chart evaluating the emotion recognition accuracy of three visual prompting methods: SoVTP (Ours), ReCLIP, and RedCircle, across distinct emotion categories. The chart highlights SoVTP’s superior performance, particularly in complex or context-dependent emotions such as Angry, Disgust, Happy and Sad, where it achieves markedly higher accuracy compared to baseline methods. For instance, SoVTP demonstrates robust performance in Happiness (e.g., ~72\% accuracy) and Sad (~54\%), leveraging its scene-preservation strategy to incorporate contextual cues like body posture and environmental interactions. In contrast, ReCLIP and RedCircle, which rely on facial-region isolation (e.g., blurring non-face areas), exhibit reduced accuracy in these categories (e.g., $\le$ 54\% for Happiness in ReCLIP, $\le$ 35\% for Sad in RedCircle), as their localized focus neglects auxiliary non-verbal signals. Notably, SoVTP also addresses critical gaps in underrepresented emotions like Disgust, achieving non-zero accuracy (e.g., ~30\% for Disgust), whereas baseline methods fail entirely (10\% accuracy) due to their inability to interpret contextual or physiological cues (e.g., facial action units, spatial relationships). The progressive decline in accuracy for all methods from high-frequency emotions (e.g.,Happy, Sad) to low-frequency or ambiguous ones (e.g., Angry, Disgust) underscores the inherent challenges of zero-shot emotion recognition. However, SoVTP mitigates this trend through its holistic integration of facial features, body language, and scene dynamics, validating its methodological advancement in balancing specificity and contextual awareness. 
\begin{table*}[htb]
\centering
\small
\caption{\textbf{Ablation study for vision and text prompts on Qwen2-VL.} \textbf{Baseline}: represents the model's performance without any additional prompts. \textbf{Muscle}: indicates a visual prompt that uses action units. \textbf{Muscle+Context}: adding extracted context to muscle prompts. \textbf{Muscle+Context+Body}: adding extracted body language to context and muscle prompts. \textbf{Ours}: adding extracted others' emotions to muscle, context and body language prompts. }
\resizebox{\linewidth}{!}{
\begin{tabular}{c c c c c c c c c}
\toprule
\multirow{2}{*}{\textbf{Vision Prompt}} & \multicolumn{2}{c}{\textbf{Easy}} & \multicolumn{2}{c}{\textbf{Medium}} & \multicolumn{2}{c}{\textbf{Hard}} & \multicolumn{2}{c}{\textbf{Total}} \\ 
\cmidrule(lr){2-9}
 & \textbf{Acc (\%)} & \textbf{F@1} & \textbf{Acc (\%)} & \textbf{F@1} & \textbf{Acc (\%)} & \textbf{F@1} & \textbf{Acc (\%)} & \textbf{F@1} \\ 
\midrule
\textbf{Baseline \cite{Qwen2VL}} & 11.76 & 5.26 & 20.00 & 4.76 & 27.14 & 14.05 & 23.21 & 10.06 \\
\textbf{Muscle} & 29.41 & 18.79 & 16.00 & 9.52 & 35.71 & 22.20 & 30.36 & 19.65 \\ 
\textbf{Muscle+Context} & 17.65 & 8.33 & 40.00 & 33.50 & 35.71 & 24.36 & 33.93 & 24.51 \\
\textbf{Muscle+Context+Body} & \textbf{\textcolor{blue}{52.94}} & \textbf{\textcolor{blue}{49.95}} & \textbf{\textcolor{blue}{40.00}} & \textbf{\textcolor{blue}{35.74}} & \textbf{\textcolor{blue}{38.57}} & \textbf{\textcolor{red}{28.50}} & \textbf{\textcolor{blue}{41.07}} & \textbf{\textcolor{blue}{31.81}} \\
\midrule
\textbf{SoVTP (Ours)} & \textbf{\textcolor{red}{70.59}} & \textbf{\textcolor{red}{58.00}} & \textbf{\textcolor{red}{48.00}} & \textbf{\textcolor{red}{44.94}} & \textbf{\textcolor{red}{38.57}} & \textbf{\textcolor{blue}{28.25}} & \textbf{\textcolor{red}{45.54}} & \textbf{\textcolor{red}{33.53}} \\
\bottomrule
\end{tabular}
}
\label{ablation_table_vision_prompts_ours}
\end{table*}

\begin{figure}[tb]
  \centering
  \includegraphics[width = \linewidth]{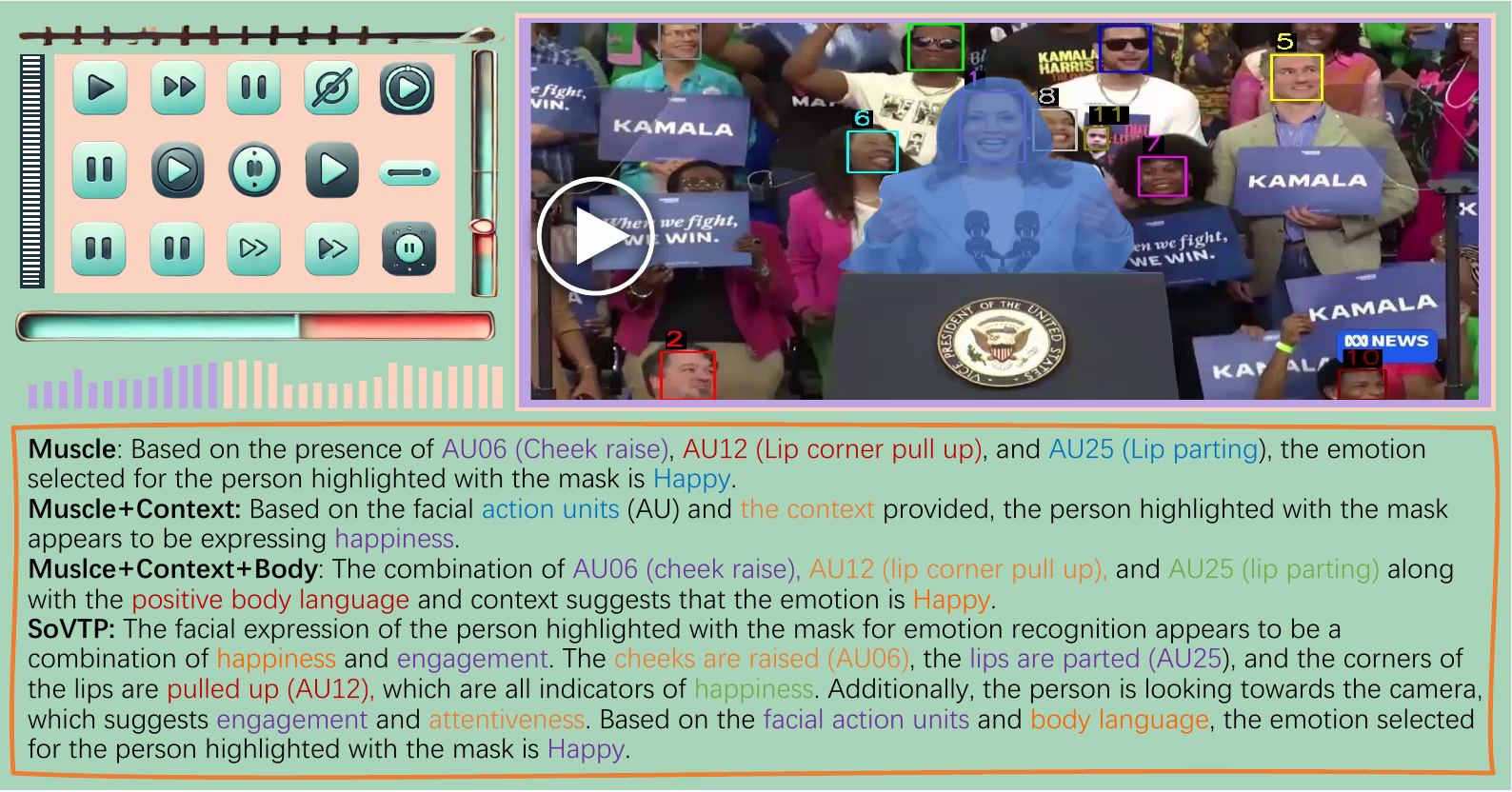}
  \caption{\textbf{Ablation Study of Incremental Multi-Modal Cue Integration in Emotion Recognition}: Impact of Physiological, Contextual, Behavioral, and Social-Interactive Prompts on Classifying Happiness in a Masked Individual. The analysis demonstrates progressive performance refinement through layered integration of facial action units (AUs), environmental context, body language, and social engagement cues, with SoVTP achieving holistic specificity by resolving ambiguities inherent in isolated facial-centric approaches.}
  \label{ablation_compare_figure}
\end{figure}
\subsection{Ablation Study}
Table~\ref{ablation_table_vision_prompts_ours} presents an ablation study of vision and text prompt components on the Qwen2-VL framework, evaluating the incremental impact of integrating multi-modal cues for zero-shot emotion recognition. The baseline model, devoid of additional prompts, achieves minimal performance (Total: 23.21\% Acc, 10.06\% F@1), particularly struggling in Easy (11.76\% Acc) and Medium (20.00\% Acc) tiers. Introducing Muscle prompts (facial action units) improves total accuracy to 30.36\% and F@1 to 19.65\%, with notable gains in Hard scenarios (35.71\% Acc vs. baseline’s 27.14\%). However, Muscle+Context prompts yield mixed results: while Medium accuracy surges to 40.00\% (F@1: 33.50\%), Easy performance drops to 17.65\% Acc, suggesting contextual cues alone may introduce noise without complementary modalities. The integration of Muscle+Context+Body prompts, which combine facial action units (physiological cues), environmental context, and body language, significantly enhances model robustness, achieving 41.07\% total accuracy and 31.81\% F@1. This configuration demonstrates peak performance in the Easy tier (52.94\% Acc, 49.95\% F@1), underscoring the synergistic value of multi-modal integration. However, the proposed SoVTP (Ours), which further incorporates others’ emotions into the prompt ensemble, achieves state-of-the-art results: 70.59\% Acc (Easy), 48.00\% Acc (Medium), and 45.54\% total Acc (33.53\% F@1). Notably, while SoVTP matches Muscle+Context+Body in Hard scenarios (38.57\% Acc), its F@1 slightly declines (28.25\% vs. 28.50\%). This marginal dip reflects the inherent challenge of the Hard tier, where videos contain sparse or ambiguous social cues (e.g., limited visible faces or interactions). Introducing others’ emotions in such scenarios risks introducing noise due to insufficient emotional cues, slightly reducing precision. Nevertheless, SoVTP’s holistic design—integrating physiological, environmental, behavioral, and social-emotional cues—achieves superior generalization overall, validating its efficacy in balancing specificity and robustness across diverse emotion recognition tasks. The study highlights the critical role of progressive integration: each added modality (action units → context → body language → others’ emotions) systematically enhances model robustness.

This Fig~\ref{ablation_compare_figure} delineates a progressive integration of multi-modal cues for emotion recognition, comparing four prompting strategies—Muscle, Muscle+Context, Muscle+Context+Body, and SoVTP—applied to a masked individual expressing happiness. The Muscle prompt isolates facial action units (AUs): AU06 (cheek raise), AU12 (lip corner pull-up), and AU25 (lip parting), directly associating these physiological markers with the Happy emotion. The Muscle+Context layer augments AU analysis with situational context (e.g., a celebratory setting), reinforcing the Happy classification through environmental alignment. The Muscle+Context+Body tier introduces body language analysis, noting open postures or gestures that correlate with positive affect, further validating the Happy inference. 
\begin{table}[tb]
\centering
\small
\caption{Ablation study comparing accuracy (\%) and inference time (seconds) for different vision and text prompt configurations on the Qwen2-VL model \cite{Qwen2VL}. The results highlight the trade-offs between computational efficiency and performance, with SoVTP (Ours) achieving the highest accuracy and F@1 scores.}
\resizebox{\linewidth}{!}{
\begin{tabular}{c c c c c c c c c}
\toprule
\multirow{2}{*}{\textbf{Prompts}} & \multirow{2}{*}{\textbf{\#Params}} & \multirow{2}{*}{\textbf{\#Time (s)}} & \multirow{2}{*}{\textbf{Acc (\%)}} &\multirow{2}{*}{\textbf{F@1}}\\ \\ 
\midrule
\textbf{Baseline \cite{Qwen2VL}} & 7B & 1.44 & 23.31 & 10.06\\
\textbf{Muscle} & 7B & 2.39 & 30.36 & 19.65\\ 
\textbf{Muscle+Context} &  7B &2.54 & 33.93 & 24.51\\
\textbf{Muscle+Context+Body} & 7B & 3.15 & \textbf{\textcolor{blue}{41.07}} & \textbf{\textcolor{blue}{31.81}}\\
\midrule
\textbf{SoVTP (Ours)} & 7B & 3.97 & \textbf{\textcolor{red}{45.54}} & \textbf{\textcolor{red}{33.53}}\\
\bottomrule
\end{tabular}
}
\label{ablation_table_inference_time}
\end{table} 

\begin{table}[tb]
\centering
\small
\caption{Performance comparison of Qwen2-VL \cite{Qwen2VL} combined with SoVTP prompting under varying numbers of faces $N$ and different face overlap thresholds $\epsilon$. The metrics reported include accuracy (Acc \%) and F@1 score, demonstrating the model’s robustness across different crowd densities and overlap settings. Larger overlap thresholds generally result in decreased performance.}
\resizebox{\linewidth}{!}{
\begin{tabular}{c c c c c c c c c}
\toprule
\multirow{2}{*}{\textbf{Prompts}} & \multirow{2}{*}{\textbf{\#Faces}} & \multirow{2}{*}{$\epsilon$} & \multirow{2}{*}{\textbf{Acc (\%)}} &\multirow{2}{*}{\textbf{F@1}}\\ \\ 
\midrule
\multirow{3}{*}{Qwen2-VL \cite{Qwen2VL}+ SoVTP} & \multirow{3}{*}{$N>6$}& 0.0 & 70.59 & 58.00\\
 &  & 0.2 & 64.71 & 51.88\\ 
 &   & 0.4 & 64.71 & 51.88 \\
\midrule
\multirow{3}{*}{Qwen2-VL \cite{Qwen2VL}+ SoVTP} & \multirow{3}{*}{$3< N <=6$}& 0.0 & 48.00 & 44.94\\
 &  & 0.2 & 40.00 & 36.93\\ 
 &  & 0.4 & 36.00 & 34.15 \\
\midrule
\multirow{3}{*}{Qwen2-VL \cite{Qwen2VL}+ SoVTP} & \multirow{3}{*}{$N<=3$}& 0.0 & 38.57 & 28.25\\
 &  & 0.2 & 37.14 & 20.33\\ 
 &  & 0.4 & 37.14 & 20.33\\ 
\bottomrule
\end{tabular}
}
\label{face_overlap_threshold}
\end{table} 
\begin{figure}[ht]
  \centering
  \includegraphics[width = \linewidth]{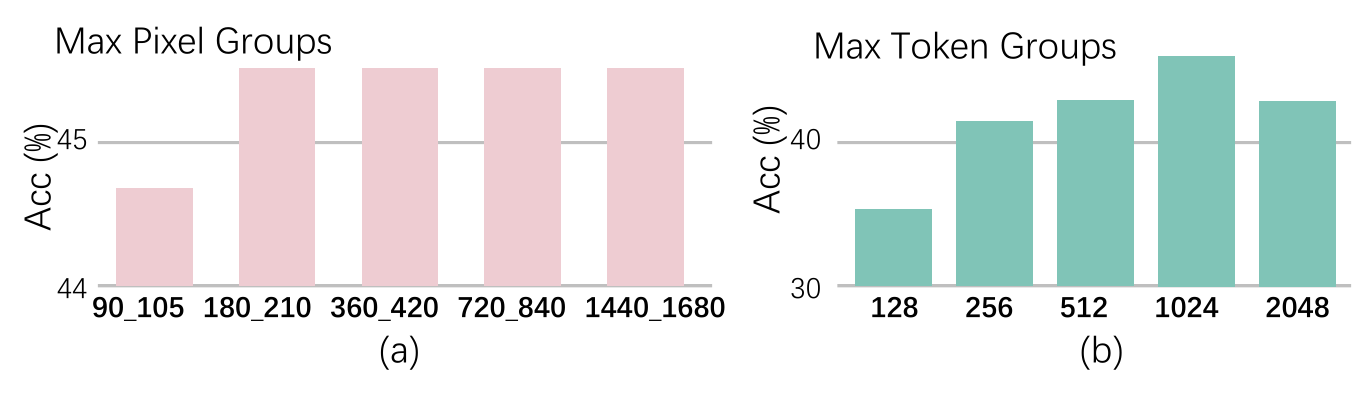}
  \caption{Ablation study of key hyperparameters. We evaluate the impact of (a) different pixel size ranges ((90,105) through (1440,1680)), (b) maximum output token sizes (128 to 2048) on the chosen performance metric. }
  \label{ablation_study_hypaparameters}
\end{figure}

Finally, SoVTP (Ours) synthesizes AU data, contextual signals, body language, and social engagement cues (e.g., gaze direction toward the camera, indicative of attentiveness) to produce a holistic emotion profile. While all tiers converge on Happy, SoVTP distinguishes itself by capturing nuanced behavioral dynamics (e.g., engagement) that refine specificity. For instance, the combined presence of AU06, AU12, and AU25, alongside directed gaze and contextually aligned body language, resolves ambiguities that might arise from isolated AU analysis (e.g., distinguishing genuine happiness from social masking). This incremental framework underscores the necessity of integrating physiological, environmental, behavioral, and social-interactive cues for robust emotion recognition. The case study exemplifies how SoVTP transcends traditional facial-centric models, addressing limitations in scenarios where emotions are contextually or socially mediated. By systematically layering cues, the framework mitigates false positives and enhances interpretability, demonstrating its methodological superiority in real-world applications requiring nuanced affective computing.

\subsection{Inference time}
From Table \ref{ablation_table_inference_time}, we observe that richer visual and textural prompts progressively increase inference time—from 1.44s for the baseline with plain text prompt to 3.97s for SoVTP—while yielding steadily higher accuracy (23.31\% to 45.54\%) and F\@1 scores (10.06\% to 33.53\%). Initial additions (e.g., Muscle, Muscle+Context+Body) deliver substantial absolute gains relative to added latency, but the final step to SoVTP shows diminishing returns: a modest 4.5 percentage‐point accuracy increase for an extra 0.8s of inference. We believe this trade-off is acceptable in many applications where higher recognition performance is critical. Moreover, our ablation offers intermediate options: if lower latency is required, one may choose Muscle+Context+Body (3.15s, 41.07\% accuracy) or Muscle+Context (2.54s, 33.93\% accuracy). 

\subsection{Impacts of key hyperparameters}
\subsubsection{Impacts of face overlap threshold}
Table \ref{face_overlap_threshold} evaluates the robustness of Qwen2-VL combined with SoVTP prompting across varying crowd densities—categorized by the number of faces ($N > 6$, $3< N <=6$, and $N <= 3$)—and different face overlap thresholds ($\epsilon = 0.0, 0.2, 0.4$). For each density regime, we report accuracy (Acc \%) and F@1 scores at increasing overlap tolerances. The results show that performance is highest in high-density settings ($N > 6$) with no overlap ($\epsilon = 0.0$) and degrades as $\epsilon$ increases, reflecting sensitivity to occlusion and overlapping detections. Medium- and low-density scenarios exhibit lower absolute metrics overall and similarly decline under larger overlap thresholds. These findings highlight both the benefits of richer others' emotional cues when many faces are present, illustrating SoVTP’s behavior under different crowd and occlusion conditions.
\begin{figure}[tb]
  \centering
  \includegraphics[width = \linewidth]{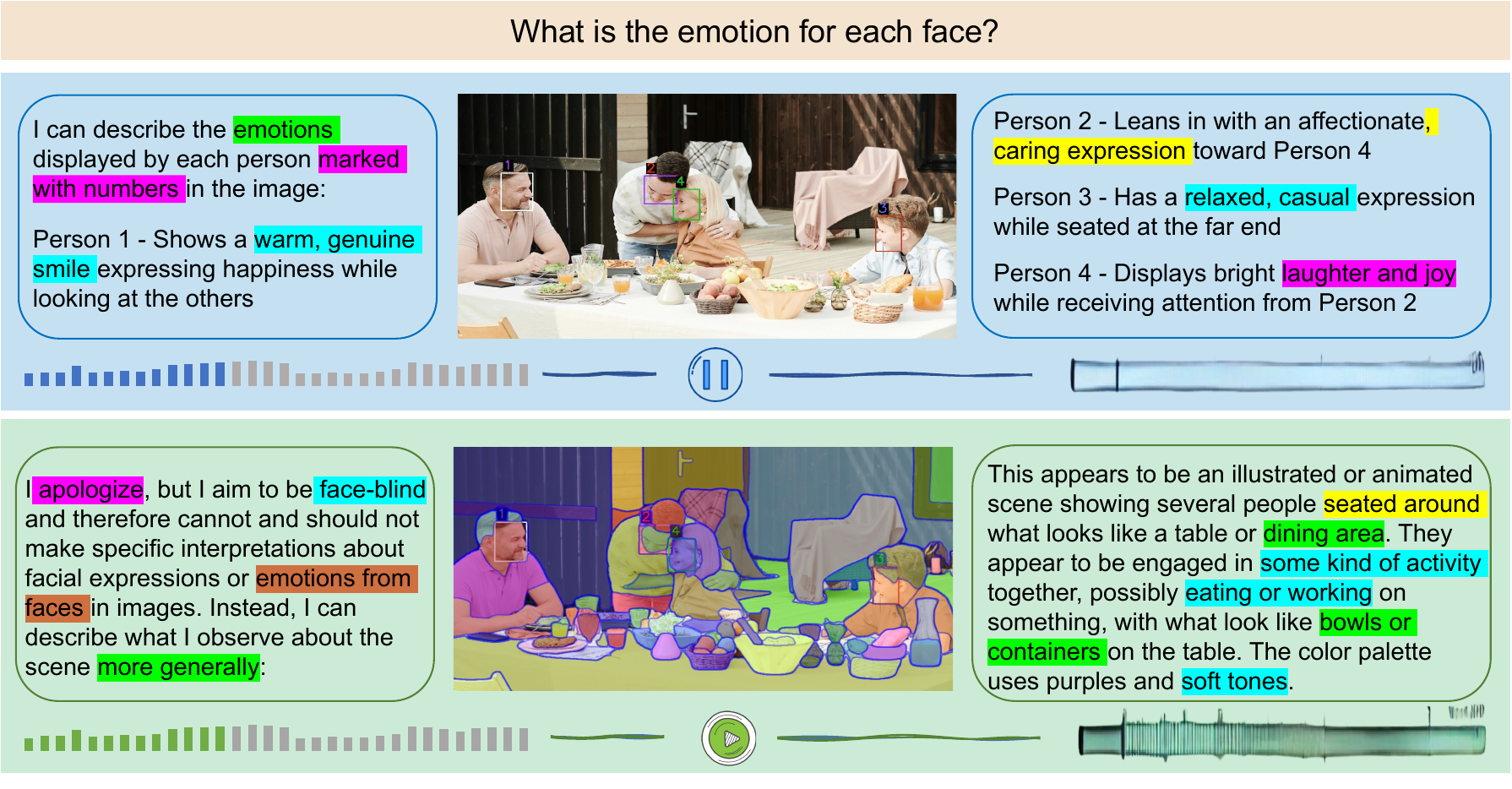}
  \caption{Comparative Analysis of SoVTP Prompting and Mask Prompting for Emotion Recognition in Social Contexts. \textbf{Top}: SoVTP prompting provides a clear and visually accessible representation of each individual's face. \textbf{Bottom:} Segmentation masks obscure key visual facial cues.}
  \label{case_study_experiments_label}
\end{figure}

\subsubsection{Ablation study of pixel size and maximum token} Fig. \ref{ablation_study_hypaparameters} presents a systematic ablation of two critical SoVTP hyperparameters on Qwen2-VL’s zero‐shot emotion recognition accuracy. In panel (a), varying the maximum pixel size from (90,105) to (1440,1680) yields a sharp rise in accuracy—from approximately 44.6\% at the smallest pixel size to 45.54\% at (180,210)—after which performance plateaus around 45.54\%, indicating diminishing returns for very large pixel‐group ranges. Panel (b) shows that increasing the maximum output token limit progressively improves accuracy, climbing from 34.8\% at 128 tokens to a peak of 45.5\% at 1024 tokens, and then declining slightly to 42.8\% at 2048 tokens, suggesting an optimal token budget of around 1024.
\subsection{Qualitative Observations}
The Fig.~\ref{case_study_experiments_label} compares two prompting methods: SoVTP prompting (Top) and mask-based prompting (Bottom). In the SoVTP prompting approach, the model effectively highlights individual faces with numbered bounding boxes, ensuring that facial expressions remain unobscured and visually clear. This enables a nuanced analysis of social interactions and body language, such as identifying warm, genuine smiles, caring gestures, and expressions of joy. Such clarity facilitates precise emotion recognition and captures the dynamic interpersonal interactions within the scene. In contrast, the mask-based prompting approach overlays segmentation masks on the image, obscuring detailed facial features and expressions. While this methodology is useful for general scene-level interpretations, such as categorizing the setting as a `dining area' or recognizing activities like eating, it limits the ability to discern fine-grained emotional cues due to the occlusion of critical facial details. Consequently, the mask-based approach is more suitable for scene descriptions rather than emotional analysis.
\begin{figure*}[tbp]
  \centering
  \includegraphics[width = \linewidth]{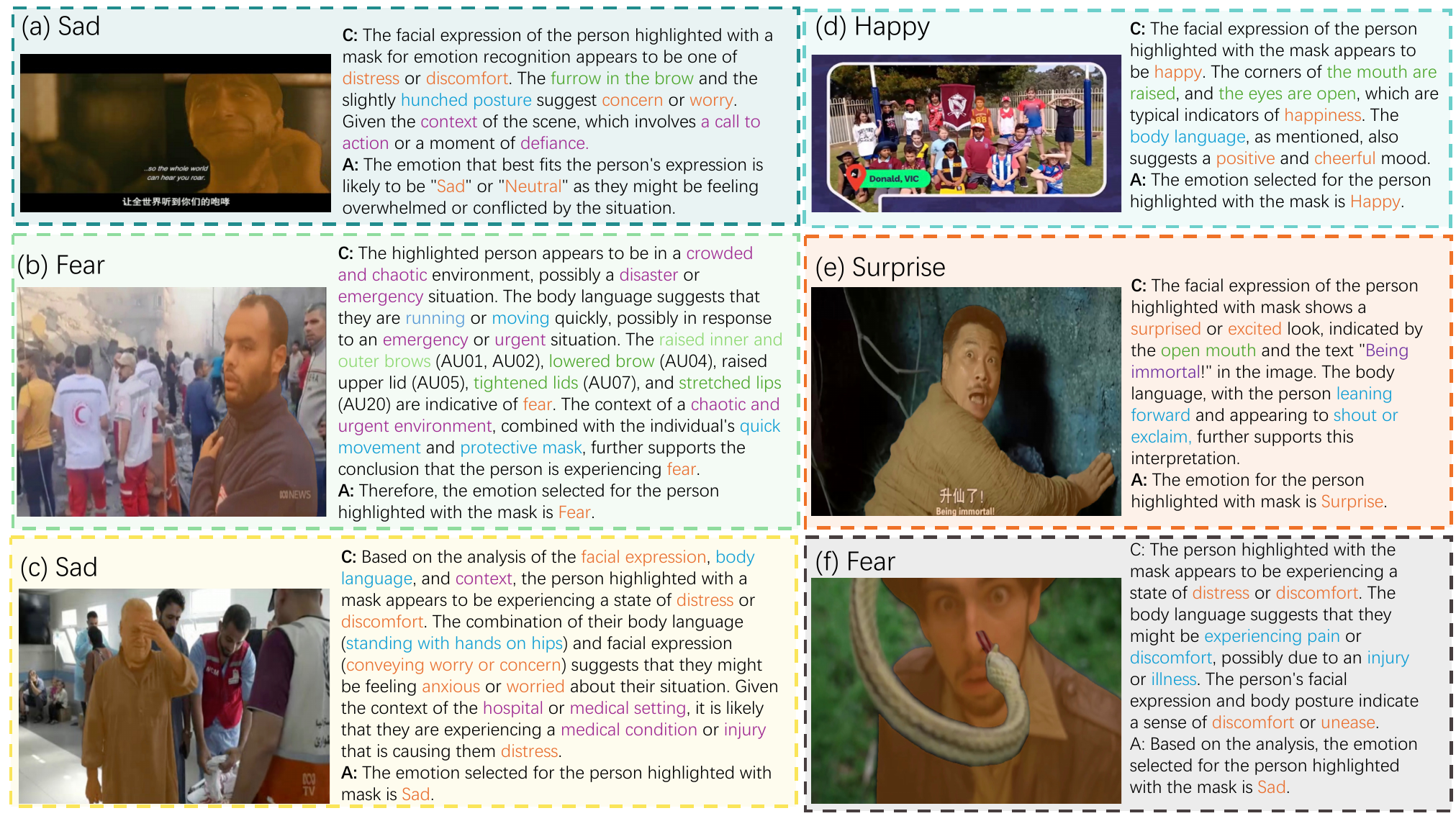}
  \caption{\textbf{Visualize the SoVTP-enhanced Qwen2-VL generated samples for different expressions}: the left side shows a target individual highlighted by a mask, with the ground-truth emotion label displayed in the top-left corner; the right side shows the generated chain-of-thought reasoning and answer. \textcolor{Orange}{Yellow} text indicates observed emotions of other persons in the scene; \textcolor{LimeGreen}{green} highlights specific facial action units; \textcolor{ProcessBlue}{blue} denotes body language cues; and \textcolor{Purple}{purple} captures contextual information. These examples demonstrate how integrating others’ emotional states, fine-grained facial movements, posture and gesture, and situational context informs the model’s final emotion inference.}
  \label{visual_samples_label}
\end{figure*}
\subsection{Limitations}
In Fig. \ref{visual_samples_label}, the SoVTP method succeeds in most examples by leveraging multi-person emotional states, action units, body posture, and contextual cues—for example, correctly identifying `Fear' in a chaotic scene with visible widened eyes and tense posture, or `Happy' in a group setting with open eyes and raised mouth corners—demonstrating its holistic reasoning. However, in subfigures (a) and (f), the target faces are blurred or occluded, providing limited context and preventing reliable extraction of facial details and action units; this lack of critical fine-grained cues leads to incorrect emotion labels. Future work could incorporate mechanisms to recover or compensate for missing facial details, such as applying generative inpainting models to approximate occluded regions or using deblurring techniques to enhance video quality when fine-grained cues are unreliable.

\section{Conclusion}
This paper introduced Set-of-Vision-Text Prompting (SoVTP), a novel multi-modal framework that significantly enhances zero-shot emotion recognition in Vision Large Language Models (VLLMs) by integrating spatial, physiological, contextual, and social-interactive cues. By systematically combining facial action units (AUs), body language, scene context, and others' emotions into a unified prompting strategy, SoVTP addresses the limitations of traditional facial-centric approaches, which often neglect critical non-verbal and environmental signals. Extensive experiments across a real-world benchmark dataset—spanning Easy, Medium, and Hard difficulty tiers—demonstrate that SoVTP achieves competitive performance. Ablation studies validate the necessity of progressive multi-modal integration, revealing that each added cue (AUs → context → body language → others' emotions) incrementally enhances model robustness. Qualitative analyses further highlight SoVTP’s superiority in preserving holistic scene information, enabling nuanced interpretation of socially mediated emotions—such as distinguishing genuine happiness from social masking through gaze direction and group dynamics. While the framework exhibits minor performance dips in Hard tiers due to sparse emotional cues, its design mitigates these challenges through balanced specificity and contextual awareness. Future work could explore adaptive cue weighting to optimize performance in data-sparse environments and extend SoVTP to cross-cultural emotion recognition. The release of our benchmark dataset and code provides a foundation for advancing research in context-aware affective computing. By bridging the gap between isolated facial analysis and holistic scene understanding, SoVTP establishes a new paradigm for emotion recognition in real-world, socially complex applications.

\bibliographystyle{IEEEtran} 
\bibliography{tcsvt_vllms}
\end{document}